\documentclass[10pt,twocolumn,letterpaper]{article}

\usepackage[pagenumbers]{cvpr} 

\usepackage{graphicx}
\usepackage{amsmath}
\usepackage{amssymb}
\usepackage{booktabs}

\usepackage[pagebackref,breaklinks,colorlinks]{hyperref}

\usepackage[capitalize]{cleveref}
\crefname{section}{Sec.}{Secs.}
\Crefname{section}{Section}{Sections}
\Crefname{table}{Table}{Tables}
\crefname{table}{Tab.}{Tabs.}

\def\cvprPaperID{39} 
\def\confName{CVPR}
\def\confYear{2022}

\usepackage{overpic}
\usepackage{enumitem} 
\usepackage{overpic} 
\usepackage{color}

\definecolor{turquoise}{cmyk}{0.65,0,0.1,0.3}
\definecolor{purple}{rgb}{0.65,0,0.65}
\definecolor{dark_green}{rgb}{0, 0.5, 0}
\definecolor{orange}{rgb}{0.8, 0.6, 0.2}
\definecolor{red}{rgb}{0.8, 0.2, 0.2}
\definecolor{darkred}{rgb}{0.6, 0.1, 0.05}
\definecolor{blueish}{rgb}{0.0, 0.3, .6}
\definecolor{light_gray}{rgb}{0.7, 0.7, .7}
\definecolor{pink}{rgb}{1, 0, 1}
\definecolor{greyblue}{rgb}{0.25, 0.25, 1}

\usepackage{blindtext}

\renewcommand{\paragraph}[1]{\vspace{1em}\noindent\textbf{#1}.}
\begin{document}
\title{Patch-wise Contrastive Style Learning for Instagram Filter Removal}

\author{Furkan Kınlı\textsuperscript{1} \qquad Barış Özcan\textsuperscript{2} \qquad Furkan Kıraç\textsuperscript{3} \\
Video, Vision and Graphics Lab\\ Özyeğin University\\
{\tt\small \{furkan.kinli\textsuperscript{1}, furkan.kirac\textsuperscript{3}\}@ozyegin.edu.tr, baris.ozcan.10097@ozu.edu.tr\textsuperscript{2}}

}
\maketitle
\begin{abstract}
Image-level corruptions and perturbations degrade the performance of CNNs on different downstream vision tasks. Social media filters are one of the most common resources of various corruptions and perturbations for real-world visual analysis applications. The negative effects of these distractive factors can be alleviated by recovering the original images with their pure style for the inference of the downstream vision tasks. Assuming these filters substantially inject a piece of additional style information to the social media images, we can formulate the problem of recovering the original versions as a reverse style transfer problem. We introduce Contrastive Instagram Filter Removal Network (CIFR), which enhances this idea for Instagram filter removal by employing a novel multi-layer patch-wise contrastive style learning mechanism. Experiments show our proposed strategy produces better qualitative and quantitative results than the previous studies. Moreover, we present the results of our additional experiments for proposed architecture within different settings. Finally, we present the inference outputs and quantitative comparison of filtered and recovered images on localization and segmentation tasks to encourage the main motivation for this problem.
\end{abstract}
\section{Introduction}
\label{sec:intro}
Social media filters (\eg Instagram filters) transform an image into a different version by applying several transformations, and this modified version may have color-level or pixel-level corruptions and perturbations. These filters modify the original image by adjusting the contrast and brightness, or changing hue and saturation, or introducing different levels of blur and noise, or applying color curves or vignetting. Though these filters convert images to a more aesthetically pleasing appearance, they also make the content in those images more complicated to understand by learning-based algorithms. Therefore, removing the filters from social media images is a crucial preprocessing step completed for the visual analysis of social media contents. 

Convolutional Neural Networks (CNNs) are one of the most common choices for solving different vision tasks, and there are several prominent studies that propose the fundamental solutions based on CNNs for these tasks such as classification \cite{He2015, huang2017densely, Simonyan15, 43022}, localization \cite{Lin_2017_CVPR, Redmon_2016_CVPR, NIPS2015_14bfa6bb, wang2020deep, zoph20selftraining}, segmentation \cite{Chen_2018_ECCV, 8237584, wang2020deep, zoph20selftraining}, tracking \cite{10.1007/978-3-319-48881-3_56, Bewley2016_sort, Wojke2017simple} and retrieval \cite{triplet, Koch2015SiameseNN}. However, the recent studies \cite{45818, nguyen2015deep, Xie_2017_ICCV} argue that CNNs are not robust to the image-level corruptions and perturbations for the downstream tasks, and this leads to a significant decrease on the performance regardless of the task. At this point, the filters applied to the social media images can be considered as the natural example of the image-level corruptions and perturbations, which can be frequently encountered in several real-world vision applications. As exemplified in \cite{Kinli_2021_CVPR}, CNNs do not give the exact segmentation outputs for the original image and its filtered versions, but also give intolerably inaccurate outputs for the filtered versions,  due to the different levels of corruptions and different types of perturbations caused by filtering. 

\begin{figure*}[ht]
\begin{center}
    \includegraphics[width=\linewidth]{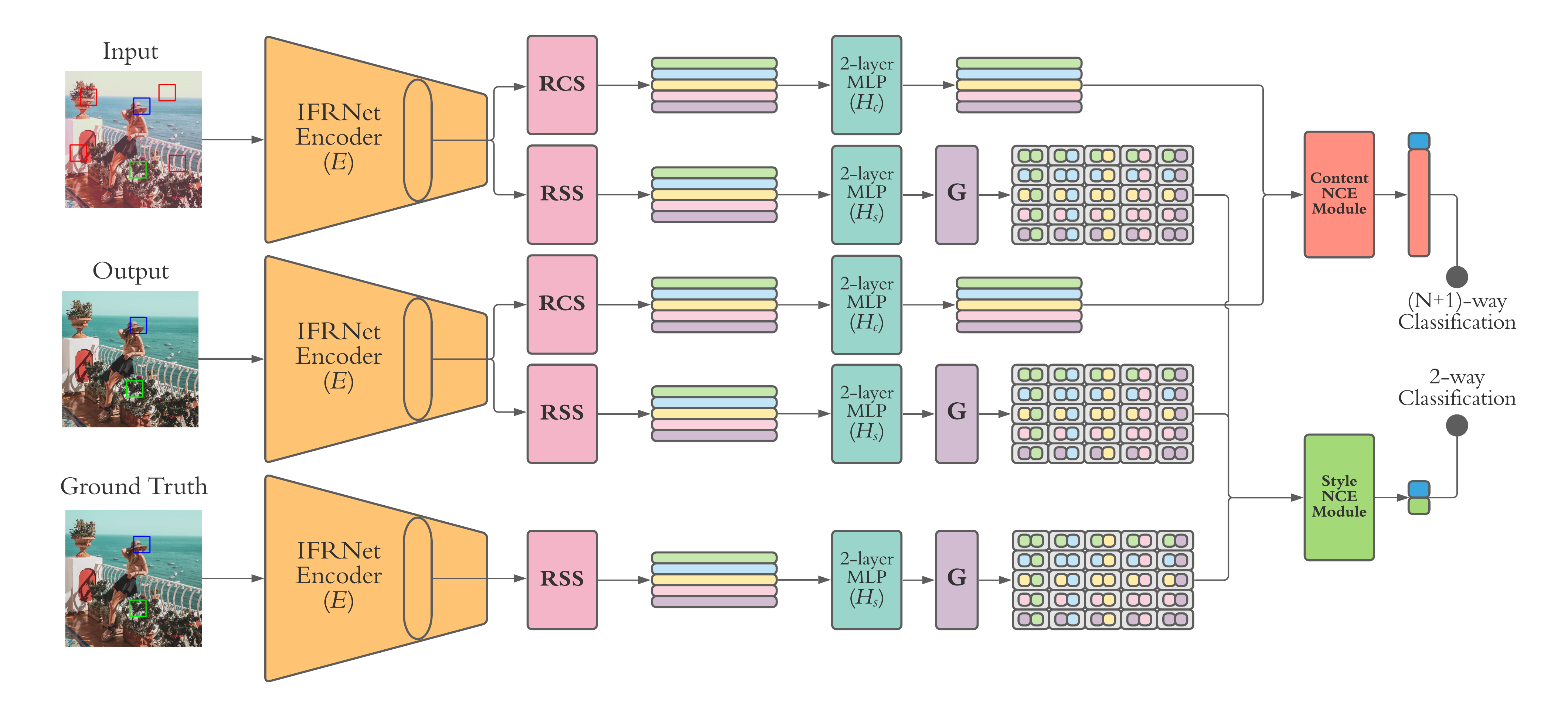}
\end{center}
   \caption{Isolated patch sampling modules for distilling the content and style information. This figure shows the pipeline for only a \textbf{single} level features. The extracted feature maps by IFRNet Encoder (\textit{$E$}) are first fed into the random sampling modules for the content (\textbf{RCS}) and style (\textbf{RSS}), separately. After encoding them by corresponding 2-layer MLP modules, (\textbf{$H_c$}) and (\textbf{$H_s$}), the \textit{content} patches for the input and the output are sent to Content NCE module, and content NCE loss is calculated as proposed in \cite{park2020cut}. Moreover, the \textit{style} patches are extracted by \textbf{G} for calculating the Gram matrix of the encoded features, and style NCE loss learns to select the patch with \textit{pure} style over the filtered patch.}
\label{fig:long}
\label{fig:onecol}
\end{figure*}

In the previous studies, there are two main approaches proposing a solution for better analyzing the filtered social media images: (1) the filter classification \cite{bianco2017artistic, filterInvariant, chu2019photo, Wu_Wu_Singh_Davis_2020}, (2) learning the transformations applied by the filter \cite{artisticFilter, photoTransform}. The main drawback for both approaches is that they do not directly try to recover the original images, but only to learn the class, or to approximate the transformation matrix of the filters applied. Recently, a novel approach for recovering the original images from the filtered versions has been proposed by \cite{Kinli_2021_CVPR}. This approach mainly assumes that the filters applied inject the additional style information to the images, and thus considering the filter removal problem as a reverse style transfer problem. We combine adaptive feature normalization idea for filter removal as in \cite{Kinli_2021_CVPR} and the patch-wise contrastive learning mechanism \cite{park2020cut}, and improve them. In this study, we propose \textit{Contrastive Instagram Filter Removal Network (CIFR)}, which employs novel patch sampling modules for contrastive semantic and style NCE losses leading to preserve the semantic information while removing the additional style information injected by the filters. This work has the following contributions:

\begin{itemize}[leftmargin=*]
\setlength\itemsep{-.3em}
\item We introduce \textit{Contrastive Instagram Filter Removal Network (CIFR)}, which enhances the idea of reverse style transfer for recovering the original images proposed in \cite{Kinli_2021_CVPR} by adding patch-wise contrastive style learning mechanism to the objective functions.
\item We compare the qualitative and quantitative results of CIFR with the benchmark presented in \cite{Kinli_2021_CVPR}. This benchmark contains the previous filter removal approaches \cite{artisticFilter, Kinli_2021_CVPR} and the fundamental \cite{pix2pix2017, CycleGAN2017} and the related \cite{FuCVPR19-GcGAN, DRIT_plus, park2020cut, sidorov2019conditional} image-to-image translation studies.
\item We present the additional results of our proposed architecture within the following settings: (1) using pre-trained weights of IFRNet \cite{Kinli_2021_CVPR}. (2) including only PatchNCE loss \cite{park2020cut} to the objective functions in \cite{Kinli_2021_CVPR}. (3) excluding Identity Regularization \cite{park2020cut} or (4) well-known consistency losses used in \cite{Kinli_2021_CVPR}.
\item We demonstrate the impact of removing the visual effects brought by Instagram filters on the performance of the downstream vision models like localization and segmentation.
\end{itemize}

\section{Related works}
\label{sec:related}
\paragraph{Instagram Filter Removal}
Removing Instagram filters is an emerging task in vision, and investigated by only limited number of studies in the literature. \cite{artisticFilter} is one of the prominent studies trying to remove the visual effects brought by Instagram filters, and it follows a strategy for adaptively learning the parametric local transformations for each filter by using CNNs. By using a similar idea, \cite{photoTransform} proposes a CNN architecture for transferring the photographic effects of a filter among the images with different contents by predicting the coefficients of the transformations applied. \cite{Kinli_2021_CVPR} introduces an adversarial methodology that directly learns to remove Instagram filters by adaptively normalizing the style information injected by filters in the feature representation of the filtered images. Moreover, there are other studies that try to recognize the filters applied to the images, instead of directly removing their effects, by using the ancestor CNNs (\eg AlexNet \cite{NIPS2012_c399862d}, LeNet \cite{Lecun98gradient-basedlearning}) \cite{bianco2017artistic}, or more commonly-used CNNs (\eg VGGs \cite{Simonyan15}, ResNets \cite{He2015}) \cite{chu2019photo} or Siamese CNNs \cite{filterInvariant}.

\paragraph{Reverse Style Transfer}
Recent studies in Style Transfer \cite{Gatys2015c, 46163, huang2017adain} demonstrate that the style information of a reference image can be transferred into a target image without losing the main context. This can be considered as \textit{many-to-many} translation \cite{pix2pix2017, attribute_hallucination, park2020cut, CycleGAN2017} where any style information is captured from the feature representation of a reference image, and then fused into the feature representation of another image. Similarly, Reverse Style Transfer is described in \cite{Kinli_2021_CVPR} as \textit{many-to-one} translation where the multiple styles injected into an image can be eliminated by adaptively normalizing its feature representations in different levels, so that we can reverse it into its pure style (\ie without any additional style injected). In this study, we mainly follow the same idea for removing Instagram filters from the images. Assuming that the filters applied to the images interpolate a particular style information to the feature maps of these images, they can be swept away from the images during the extraction of feature representations. 

\paragraph{Contrastive Learning}
Contrastive learning is one of the most popular strategies in representation learning. Recent studies \cite{pmlr-v119-chen20j, He_2020_CVPR, hjelm2019learning, Tian_2020, oord2018representation} show that a methodology of \textit{maximizing mutual information} is capable of learning more effective representations without requiring any supervision or hand-crafted objective functions. Noise contrastive estimation (NCE) \cite{pmlr-v9-gutmann10a} has become the popular choice for this purpose, and \cite{NIPS2013_db2b4182} demonstrates that NCE can learn the feature representations in a better and efficient manner. NCE basically builds on the notion of semantic similarity among the associated signals where more similar ones are represented in more similar ways. These associated signals can be an image with itself \cite{journals/pami/DosovitskiyFSRB16, He_2020_CVPR, malisiewicz-iccv11, shrivastava-sa11}, an image with its feature representation \cite{hjelm2019learning}, an image with its patches \cite{Isola2015LearningVG, oord2018representation}, or multiple views \cite{Tian_2020} or its different transformed versions \cite{pmlr-v119-chen20j}. Moreover, \cite{park2020cut} employs this mechanism for conditional image synthesis task in multi-layer and patch-wise manner. In this study, we adapt the patch-wise contrastive learning methodology proposed in \cite{park2020cut} to reverse style transfer task, and introduce the idea of using isolated patch sampling modules for the content and style information for distilling the semantic and style similarities among the signals. 

\section{Methodology}
\subsection{Patch-wise Contrastive Style Learning}
\label{sec:patch_sampling}
Following the same assumption for the definition of reverse style transferring idea in \cite{Kinli_2021_CVPR}, Instagram filter removal task can be explained in such a way that a given image $\mathbf{\Tilde{X}} \in \mathbb{R}^{H \times W \times 3}$ including a style information of an arbitrary filter injected by some transformation functions $\mathbf{T}(\cdot)$ is turned into its original version $\mathbf{X} \in \mathbb{R}^{H \times W \times 3}$ (\ie without any additional style information injected) by a style removal module $\mathbf{F}(\cdot)$. The main purpose in this task is to discover the best style removal module for given images with different non-linear transformations applied.

\begin{equation}
    \mathbf{X} = \mathbf{F}(\mathbf{\Tilde{X}})
\end{equation}
where $\mathbf{\Tilde{X}} = \mathbf{T}(\mathbf{X})$ and $\mathbf{T}(\cdot)$ is a general transformation function representing one or more transformations applied to $\mathbf{X}$. Since finding $\mathbf{T}^{-1}(\cdot)$ for each single image is an ill-posed problem, we need to discover the best possible $\mathbf{F}(\cdot)$, which can be substituted with $\mathbf{T}^{-1}(\cdot)$ with the minimum amount of reconstruction error.

Contrastive learning can be described as building the representations of the instances on the notion of semantic similarity among their associated signals. These signals can be represented as a \textit{query} with its corresponding example instance (\ie \textit{positive}) and some non-corresponding example instances (\ie \textit{negatives}). The query $\mathbf{v} \in \mathbb{R}^{K}$, the positive $\mathbf{v}^+ \in \mathbb{R}^{K}$ and $N$ negatives $\mathbf{v}^- \in \mathbb{R}^{N \times K}$ are mapped into $K$-dimensional vectors by an encoding structure. Note that these vectors are required to be unit vectors to avoid collapsing and exploding in their space, and thus should be normalized. The problem is formulated as an $(N+1)$-way classification problem to maximize mutual information, where the query and the positive instances are closer to each other, while the query is located to far away from the negatives in the vector space. The objective function for this problem stands for learning to select the positive instance over the negatives for a particular query instance, and it is defined in Equation \ref{eq:contrastive_eq}.
\begin{equation}
\resizebox{\linewidth}{!}{
    $\boldsymbol\ell(\mathbf{v}, \mathbf{v}^+, \mathbf{v}^-) = -log\left[\frac{exp(\mathbf{v} \cdot \mathbf{v}^+ / \tau)}{exp(\mathbf{v} \cdot \mathbf{v}^+ / \tau) + \sum_{n=1}^{N}{exp(\mathbf{v} \cdot \mathbf{v}^- / \tau)}}\right]$
}
\label{eq:contrastive_eq}
\end{equation}
where $N$ is the number of negative instances and $\tau$ is the temperature parameter for scaling.

In this study, we employ the multi-layer and patch-based contrastive learning objective \cite{park2020cut} to eliminate the visual effects brought by Instagram filters. A particular patch of an image filtered by any arbitrary Instagram filter should associate with the patch of its original version at the exact location. The other patches typically do not associate with this patch. The patch at each spatial location can be represented by the feature maps computed by an encoder in a different scale in each layer. Note that the feature maps in deeper layers correspond to larger patches. We can demonstrate this patch sampling module as follows:
\begin{gather}
    \{\mathbf{z^\textit{l}}\}_L = \{H^\textit{l}(E^\textit{l}(\mathbf{\Tilde{X}}))\}_L\nonumber\\
    \{\mathbf{\hat{z}^\textit{l}}\}_L = \{H^\textit{l}(E^\textit{l}(\mathbf{\hat{X}}))\}_L
\end{gather}
where $\mathbf{z^\textit{l}}$ is the feature map of filtered image in $\textit{l}$-th layer, $\mathbf{\hat{z}^\textit{l}}$ is the feature map of unfiltered image, $H^\textit{l}$ is the mapper network for patch sampling (\ie a two-layer MLP network as in \cite{pmlr-v119-chen20j}), $E^\textit{l}$ is the encoder network, $L$ is the number of layers in $E$, $\mathbf{\Tilde{X}}$ and $\mathbf{\hat{X}}$ represent the filtered and unfiltered images, respectively. 

\begin{figure*}[t]
\begin{center}
\includegraphics[width=\textwidth]{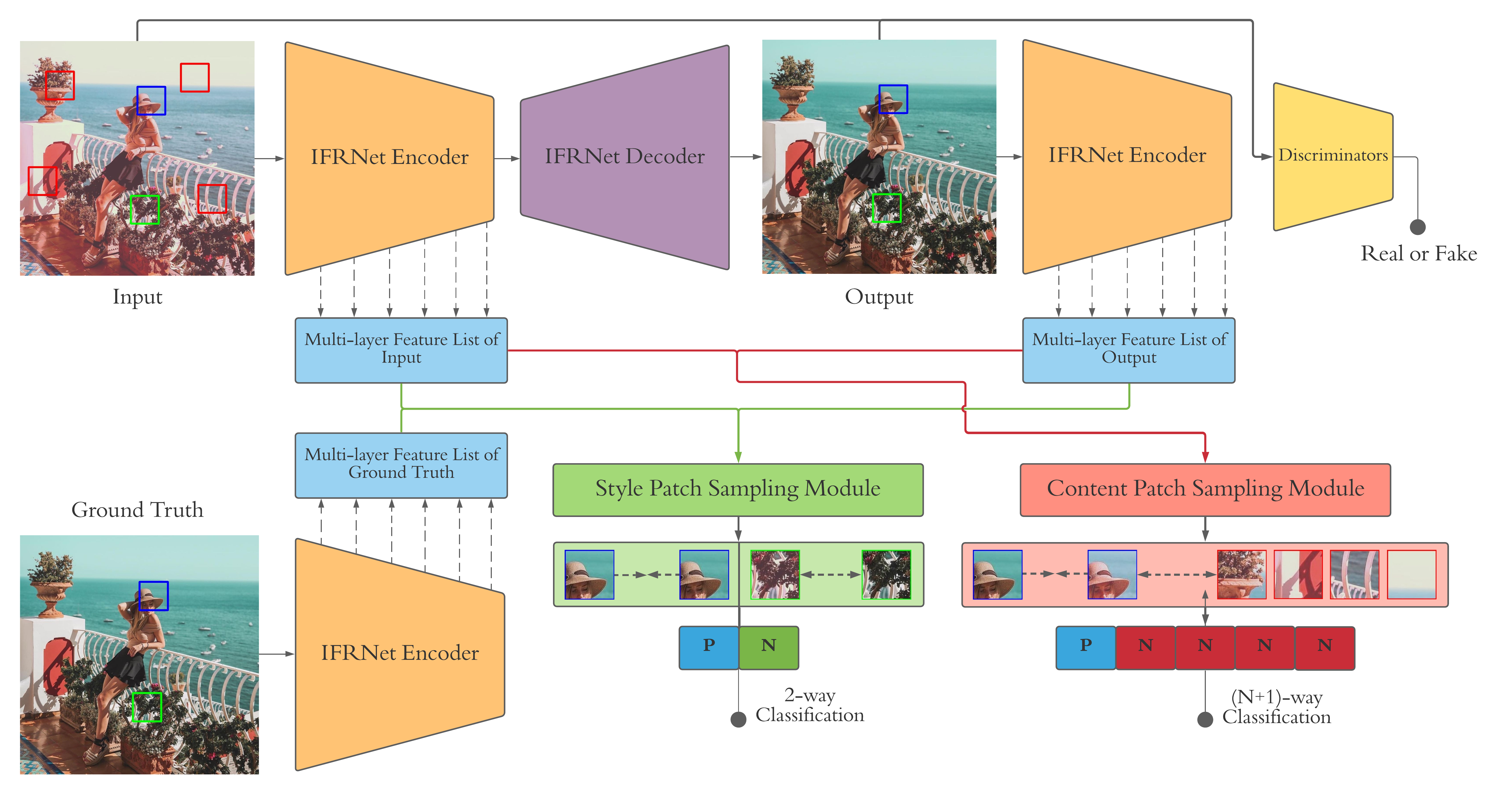}
\end{center}
   \caption{Overall architecture of Contrastive Instagram Removal Network (CIFR). The stacked features extracted by IFRNet encoder \cite{Kinli_2021_CVPR} are fed into our proposed isolated patch sampling modules. For each patch level, content NCE loss and style NCE loss are disjointly calculated for distilling the content and style information.}
\label{fig:architecture}
\end{figure*}

We extend this idea in \cite{park2020cut} by using isolated patch sampling modules for the content and style information, where the single level pipeline can be seen in Figure \ref{fig:onecol}. The main motivation behind this practice is to distill the learning process of the semantic and style similarities among the associated patches. At this point, we can leverage the original version of the images to capture the \textit{pure} style (\ie without any additional style injected). To achieve this, we extract the Gram matrices of the feature maps to express the style information via the feature correlations, and employ them to our contrastive learning pipeline. We can formulate the isolated patch sampling modules as follows:
\begin{gather}
    \{\mathbf{\Tilde{z}}_c^\textit{l},\mathbf{\Tilde{z}}_s^\textit{l}\}_L = \{H_c^\textit{l}(E^\textit{l}(\mathbf{\Tilde{X}})),\mathbf{G}^l(H_s^\textit{l}(E^\textit{l}(\mathbf{\Tilde{X}})))\}_L\nonumber\\
    \{\mathbf{\hat{z}}_c^\textit{l}, \mathbf{\hat{z}}_s^\textit{l}\}_L = \{H_c^\textit{l}(E^\textit{l}(\mathbf{\hat{X}})),\mathbf{G}^l(H_s^\textit{l}(E^\textit{l}(\mathbf{\hat{X}})))\}_L\nonumber\\
    \{\mathbf{z}_c^\textit{l},\mathbf{z}_s^\textit{l}\}_L = \{H_c^\textit{l}(E^\textit{l}(\mathbf{X})),\mathbf{G}^l(H_s^\textit{l}(E^\textit{l}(\mathbf{X})))\}_L
\end{gather}
where $\mathbf{G}^l \in \mathbb{R}^{K \times K}$ is the Gram matrix, the inner product between the features mapped by $H_s^\textit{l}$ in $l$-th layer, $\mathbf{\Tilde{z}}_c,\mathbf{\hat{z}}_c,\mathbf{z}_c$ stand for the content feature maps and $\mathbf{\Tilde{z}}_s,\mathbf{\hat{z}}_s,\mathbf{z}_s$ for the style feature maps of the filtered, unfiltered and original images, respectively. $H_c$ represents the content patch sampling module, while $H_s$ is the style patch sampling module.

For the content matching, we try to match the corresponding filtered and unfiltered patches at the same location, while exploiting the other patches within the filtered image as negatives. We also try to emulate the \textit{pure} style in the patches of the original image for the unfiltered patches. Note that, within the computational constraint, the most affordable way of capturing the pure style in such a contrastive setup is to build the strategy as 2-way classification where the negative instance has the exact same semantic information, but with different style injected. At this point, we combine two contrastive learning objectives, namely content NCE $\mathcal{L}_C$, and style NCE $\mathcal{L}_S$, for extracting the content and style information separately. Our extended version of PatchNCE loss is shown in Equation \ref{eq:ext_patchnce}.
\begin{gather}
    \mathcal{L}_{C}(E,H,\mathbf{X},\Tilde{\mathbf{X}})=\mathbb{E}_{\mathit{x{\raise.01ex\hbox{$\scriptstyle\sim$}}\mathbf{X},\Tilde{x}{\raise.01ex\hbox{$\scriptstyle\sim$}}\Tilde{\mathbf{X}}}}\sum_{l=1}^L\sum_{t=1}^{T^l}\boldsymbol\ell(\mathbf{\hat{z}}_c^\textit{l,t},\mathbf{\Tilde{z}}_c^\textit{l,t},\mathbf{\Tilde{z}}_c^\textit{l,T\textbackslash t})\nonumber\\
    \mathcal{L}_{S}(E,H,\mathbf{X},\Tilde{\mathbf{X}})=\mathbb{E}_{\mathit{x{\raise.01ex\hbox{$\scriptstyle\sim$}}\mathbf{X},\Tilde{x}{\raise.01ex\hbox{$\scriptstyle\sim$}}\Tilde{\mathbf{X}}}}\sum_{l=1}^L\sum_{t=1}^{T^l}\boldsymbol\ell(\mathbf{\hat{z}}_s^\textit{l,t},\mathbf{z}_s^\textit{l,t},\mathbf{\Tilde{z}}_s^\textit{l,t'}) \nonumber \\
    \mathcal{L}_{PatchNCE}=\gamma_c\mathcal{L}_C+\gamma_s\mathcal{L}_S
    \label{eq:ext_patchnce}
\end{gather}
where $T^l$ is the list of different spatial locations for patches at $l$-th layer, $t'$ represents a single arbitrary spatial location different than $t$, $\gamma_c$ and $\gamma_s$ are the coefficients of the content and style NCE losses. In our experiments, both coefficients are set to $0.5$.

\subsection{Architecture}

In our architecture design, we mostly follow the design of IFRNet proposed in \cite{Kinli_2021_CVPR}. IFRNet has an encoder-decoder structure with a style extractor module for applying adaptive feature normalization to all layers in the encoder. Style extractor module learns to adapt the affine parameters for the feature representations encoded by a pre-trained VGG network by using different fully-connected heads for each layer. The affine parameters are sent into adaptive instance normalization (AdaIN) \cite{huang2017adain} layers in each encoder level to eliminate the external style information. Note that any related information about the original style is retained by including skip connections to the normalized feature maps \cite{Kinli_2021_CVPR}. At the end, we stack the features in all levels in the encoder, and feed this multi-layer feature list to our proposed isolated patch sampling modules for the content and style information. We introduce the technical details about isolated patch sampling modules in Section \ref{sec:patch_sampling}.

As distinguished from IFRNet, we do not have an auxiliary classifier for classifying the filter type in our design. In \cite{Kinli_2021_CVPR}, the auxiliary classifier has been used for satisfying a naive way of maximizing mutual information between corresponding input and output instances. However, we do not prefer to include this kind of a classifier since we can achieve this practice in a more elegant way via Noise Contrastive Estimation \cite{pmlr-v9-gutmann10a}.

We feed the latent representations with no external style information to the decoder of IFRNet. The decoder contains six consecutive upsampling and residual convolutional blocks, and learns to generate the recovered image with adversarial training. Discriminators are designed as in \cite{pix2pix2017, Kinli_2021_CVPR} to penalize the global image and local patches at different scales. Our proposed architecture, namely \textit{Contrastive Instagram Filter Removal Network (CIFR)} is shown in Figure \ref{fig:architecture}.

\subsection{Objective Function}

In our study, we combine three different objective functions for our pipeline: (1) multi-layer patch-wise contrastive style loss, (2) consistency loss, (3) adversarial loss. The first one is introduced in Equation \ref{eq:ext_patchnce}, and it is responsible for distilling the learning process of the patch-level semantic and style similarities. Next, we include the consistency loss used in \cite{Kinli_2021_CVPR} to our final objective function in order to ensure the semantic and texture consistency of the output. We also employ a common adversarial training strategy (\ie \textit{WGAN-GP}) \cite{pmlr-v70-arjovsky17a} in order to enhance the realism of the recovered images, and it is demonstrated in Equation \ref{eq:adv_loss}. Our final objective function for the generator $\mathcal{L}_{G}$ can be seen in Equation \ref{eq:final_loss}.
\begin{gather}
    \mathcal{L}_{WGAN-GP}^D = -\mathbb{E}_{\mathit{x{\raise.01ex\hbox{$\scriptstyle\sim$}}\mathbf{X}}}[D(\mathit{x})] + \mathbb{E}_{\mathit{\hat{x}{\raise.01ex\hbox{$\scriptstyle\sim$}}\hat{\mathbf{X}}}}[D(\mathit{\hat{x}})] + \lambda_{gp}\mathcal{L}_{GP} \nonumber 
    \\ \mathcal{L}_{WGAN-GP}^G = -\mathbb{E}_{\mathit{\hat{x}{\raise.01ex\hbox{$\scriptstyle\sim$}}\hat{\mathbf{X}}}}[D(\mathit{\hat{x}})]
    \label{eq:adv_loss}
\end{gather}
where $D$ stands for the discriminator network, $\mathcal{L}_{GP}$ is the gradient penalty term. 
\begin{equation}
    \label{eq:final_loss}
    \mathcal{L}_{G} = \lambda_{p}\mathcal{L}_{PatchNCE} + \lambda_{c}\mathcal{L}_{Cons} + \lambda_{a}\mathcal{L}_{WGAN-GP}^G
\end{equation}
where $\lambda_{p}$, $\lambda_{c}$ and $\lambda_{a}$ are the coefficients for the patch-wise NCE loss, consistency loss and adversarial loss, and set to  $5 \times 10^{-1}$, $10^{-3}$ and $10^{-3}$, respectively.

\section{Results}
In this study, we investigate the performance of multi-layer patch-wise contrastive learning approach on Instagram filter removal task, which can be described as a reverse style transfer problem. We compare the performance of our proposed architecture, namely \textit{CIFR}, against the previous filter removal approaches \cite{artisticFilter, Kinli_2021_CVPR}, the fundamental \cite{pix2pix2017, CycleGAN2017} and the related \cite{DRIT_plus, park2020cut, FuCVPR19-GcGAN, sidorov2019conditional} image-to-image translation studies, and its own variants with different training settings. Note that we have obtained the available results in \cite{Kinli_2021_CVPR}, and re-trained the rest of compared methods on IFFI dataset with their default hyper-parameters settings. Lastly, we show the qualitative and quantitative impact of removing Instagram filters from the images on the performance of downstream vision tasks (\ie localization, segmentation).
\subsection{Experimental Setup}

We tested our methodology on IFFI dataset, which is introduced by \cite{Kinli_2021_CVPR}, and contains 9,600 high-resolution and aesthetically pleasing images along with their filtered versions by 16 different Instagram filters. There are 8,000 training and 1,600 test images in this dataset. In our experiments, we resized the images to the resolution of 256, and only applied random horizontal flipping before feeding them to our proposed model. We used the pre-trained weights of IFRNet \cite{Kinli_2021_CVPR} in our default settings. We have picked Adam optimizer \cite{DBLP:journals/corr/KingmaB14} with $\beta_1=0.5$ and $\beta_2=0.999$ for all modules, and the learning rates for the generator, discriminator and patch sampling modules are set to $2 \times 10^{-4}$, $10^{-4}$ and $10^{-5}$, respectively. We did not use any scheduling for the learning rates during training. The temperature parameter $\tau$ is set to 0.07. We have conducted our experiments on 2x NVIDIA RTX 2080Ti GPU with batch size of 8. We have trained proposed architecture for 40K steps for pre-trained settings, and 120K for training from scratch. We have implemented the code in PyTorch \cite{NEURIPS2019_9015}. The source code can be found at \url{https://github.com/birdortyedi/cifr-pytorch}.
\captionsetup[subfigure]{labelfont=bf, labelformat=empty}
\begin{figure*}[!ht]
        \centering
        \begin{subfigure}{0.095\textwidth}
                \includegraphics[width=\textwidth]{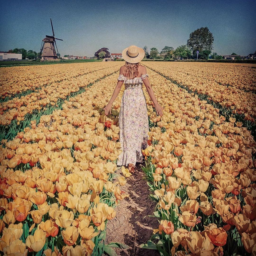}
                \includegraphics[width=\textwidth]{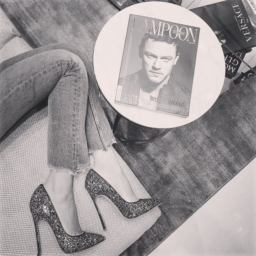}
                \includegraphics[width=\textwidth]{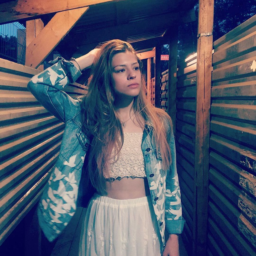}
                \includegraphics[width=\textwidth]{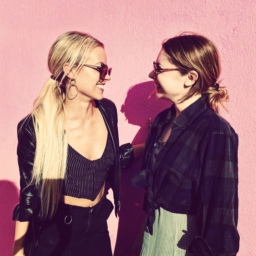}
                \includegraphics[width=\textwidth]{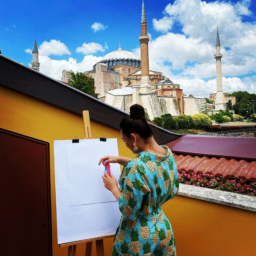}
                \includegraphics[width=\textwidth]{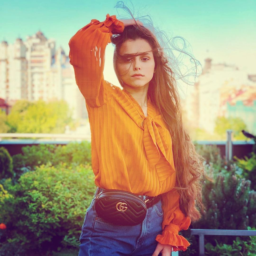}
                \caption{Filtered}
                \label{fig:fig3-filtered}
        \end{subfigure}       
        \begin{subfigure}{0.095\textwidth}
                \includegraphics[width=\textwidth]{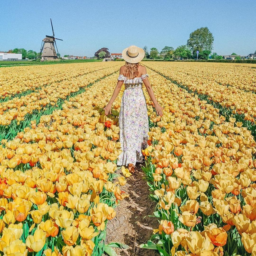}
                \includegraphics[width=\textwidth]{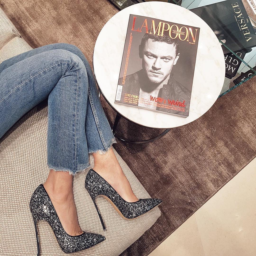}
                \includegraphics[width=\textwidth]{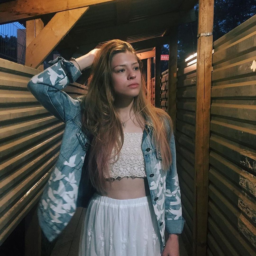}
                \includegraphics[width=\textwidth]{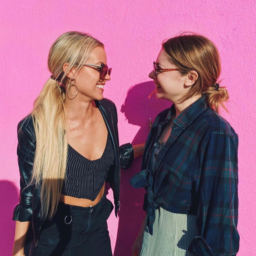}
                \includegraphics[width=\textwidth]{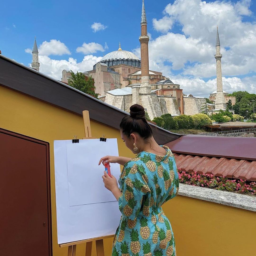}
                \includegraphics[width=\textwidth]{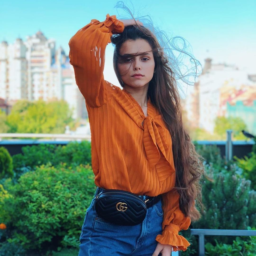}
                \caption{Original}
                \label{fig:fig3-org}
        \end{subfigure}
        \begin{subfigure}{0.095\textwidth}
                \includegraphics[width=\textwidth]{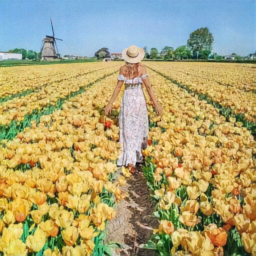}
                \includegraphics[width=\textwidth]{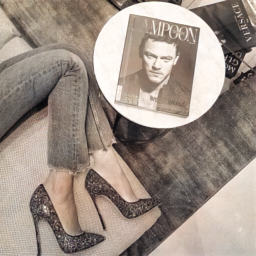}
                \includegraphics[width=\textwidth]{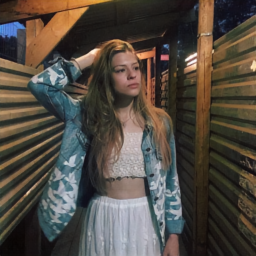}
                \includegraphics[width=\textwidth]{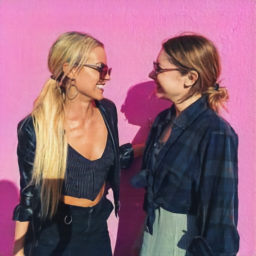}
                \includegraphics[width=\textwidth]{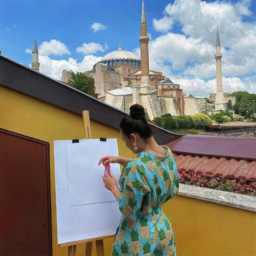}
                \includegraphics[width=\textwidth]{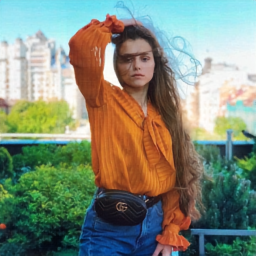}
                \caption{CIFR \textbf{(ours)}}
                \label{fig:fig3-cifr}
        \end{subfigure}
        \begin{subfigure}{0.095\textwidth}
                \includegraphics[width=\textwidth]{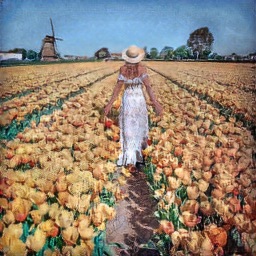}
                \includegraphics[width=\textwidth]{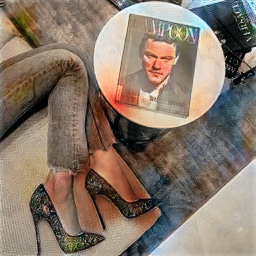}
                \includegraphics[width=\textwidth]{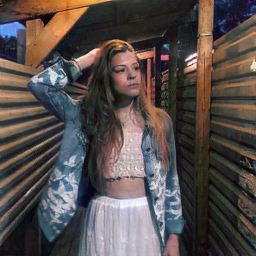}
                \includegraphics[width=\textwidth]{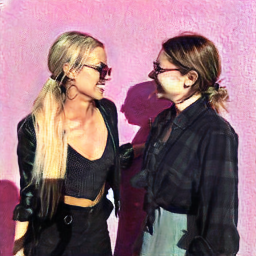}
                \includegraphics[width=\textwidth]{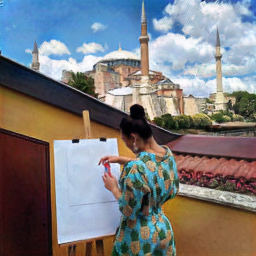}
                \includegraphics[width=\textwidth]{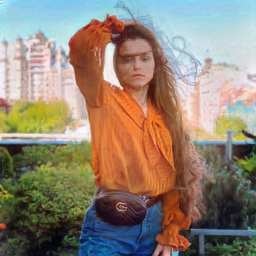}
                \caption{CUT}
                \label{fig:fig3-cut}
        \end{subfigure}
        \begin{subfigure}{0.095\textwidth}
                \includegraphics[width=\textwidth]{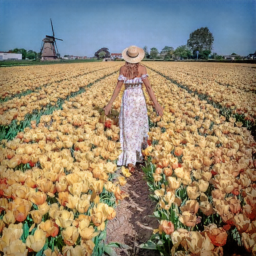}
                \includegraphics[width=\textwidth]{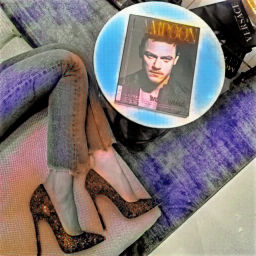}
                \includegraphics[width=\textwidth]{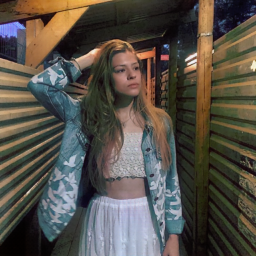}
                \includegraphics[width=\textwidth]{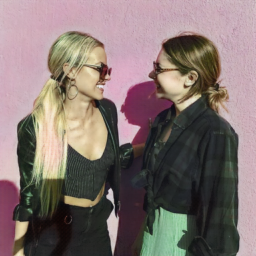}
                \includegraphics[width=\textwidth]{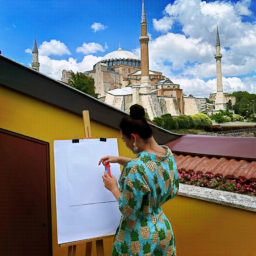}
                \includegraphics[width=\textwidth]{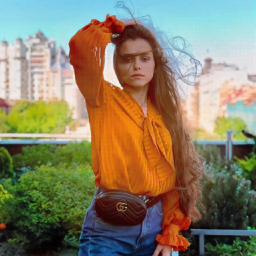}
                \caption{GcGAN}
                \label{fig:fig3-gcgan}
        \end{subfigure}
        \begin{subfigure}{0.095\textwidth}
                \includegraphics[width=\textwidth]{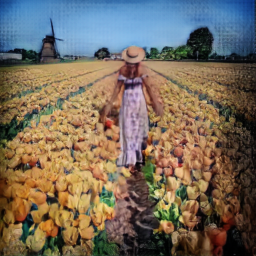}
                \includegraphics[width=\textwidth]{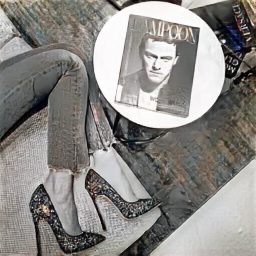}
                \includegraphics[width=\textwidth]{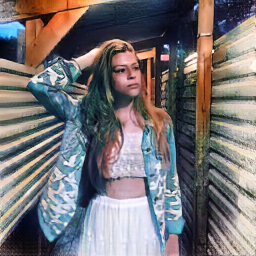}
                \includegraphics[width=\textwidth]{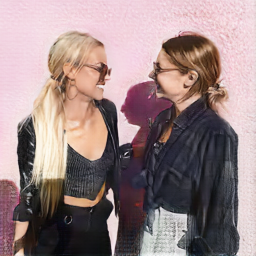}
                \includegraphics[width=\textwidth]{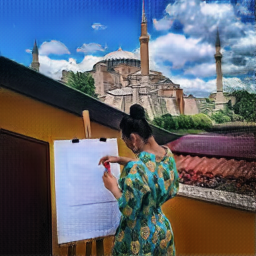}
                \includegraphics[width=\textwidth]{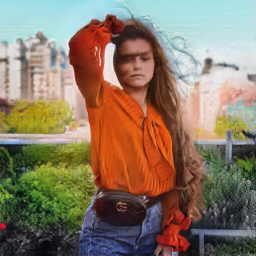}
                \caption{DRIT++}
                \label{fig:fig3-drit_plus}
        \end{subfigure}
        \begin{subfigure}{0.095\textwidth}
                \includegraphics[width=\textwidth]{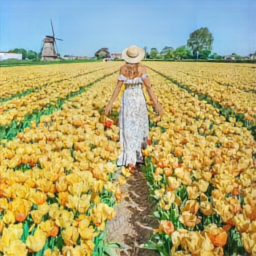}
                \includegraphics[width=\textwidth]{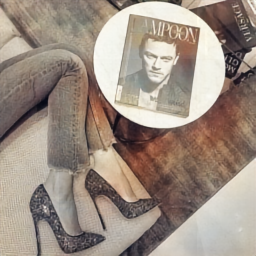}
                \includegraphics[width=\textwidth]{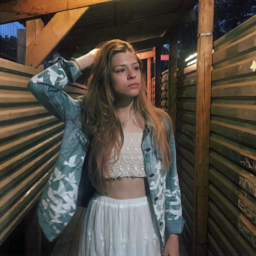}
                \includegraphics[width=\textwidth]{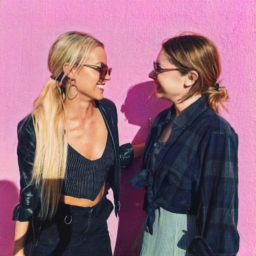}
                \includegraphics[width=\textwidth]{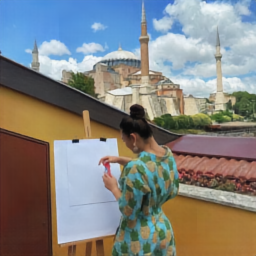}
                \includegraphics[width=\textwidth]{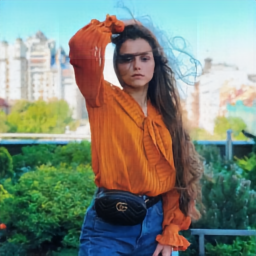}
                \caption{IFRNet}
                \label{fig:fig3-ifrnet}
        \end{subfigure}
        \begin{subfigure}{0.095\textwidth}
                \includegraphics[width=\textwidth]{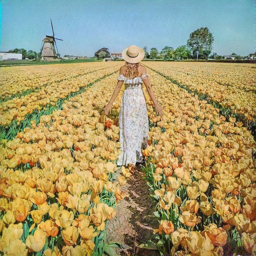}
                \includegraphics[width=\textwidth]{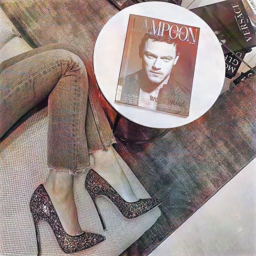}
                \includegraphics[width=\textwidth]{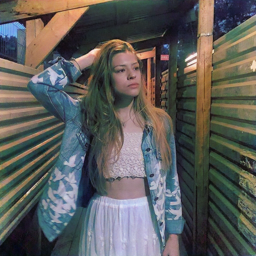}
                \includegraphics[width=\textwidth]{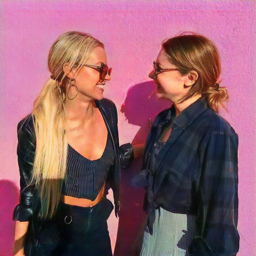}
                \includegraphics[width=\textwidth]{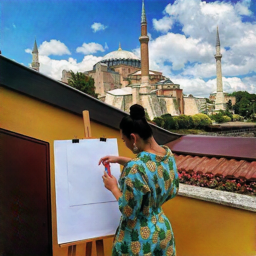}
                \includegraphics[width=\textwidth]{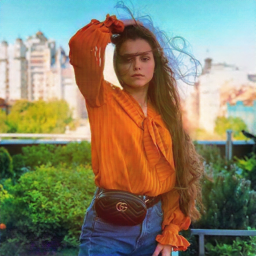}
                \caption{AngularGAN}
                \label{fig:fig3-angular}
        \end{subfigure}
        \begin{subfigure}{0.095\textwidth}
                \includegraphics[width=\textwidth]{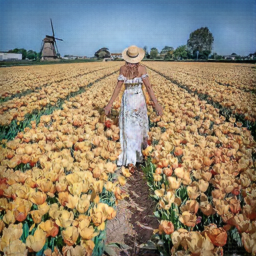}
                \includegraphics[width=\textwidth]{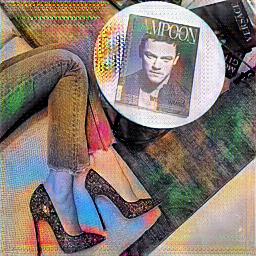}
                \includegraphics[width=\textwidth]{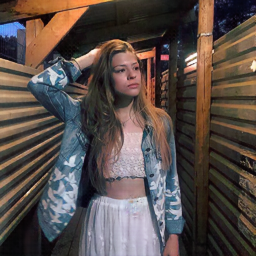}
                \includegraphics[width=\textwidth]{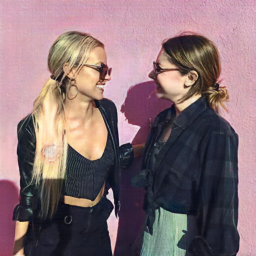}
                \includegraphics[width=\textwidth]{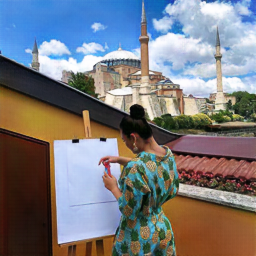}
                \includegraphics[width=\textwidth]{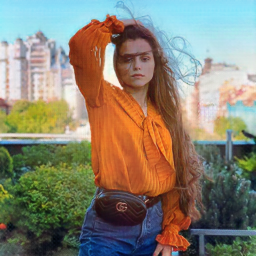}
                \caption{CycleGAN}
                \label{fig:fig3-cycle}
        \end{subfigure}
        \begin{subfigure}{0.095\textwidth}
                \includegraphics[width=\textwidth]{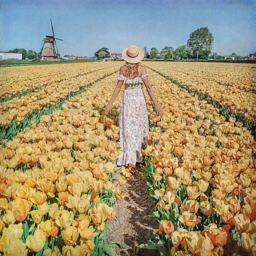}
                \includegraphics[width=\textwidth]{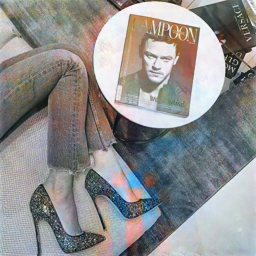}
                \includegraphics[width=\textwidth]{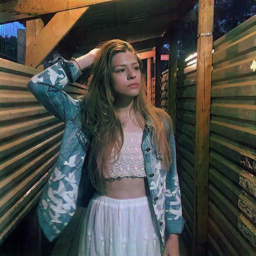}
                \includegraphics[width=\textwidth]{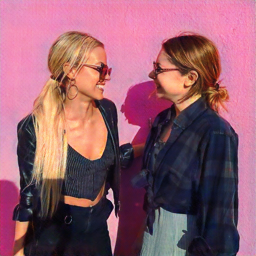}
                \includegraphics[width=\textwidth]{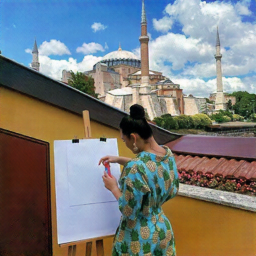}
                \includegraphics[width=\textwidth]{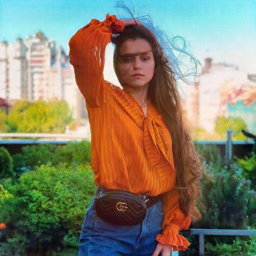}
                \caption{pix2pix}
                \label{fig:fig3-pix2pix}
        \end{subfigure}
        
        \caption{Comparison of the qualitative results of Instagram filter removal on IFFI dataset. Filters applied (top to bottom): \textit{Sutro}, \textit{Willow}, \textit{Nashville}, \textit{Amaro}, \textit{Lo-Fi}, \textit{Toaster}.}\label{fig:fig3} 
\end{figure*}

\subsection{Qualitative Comparison}

\begin{figure}[h]
        \centering
        \includegraphics[width=\linewidth]{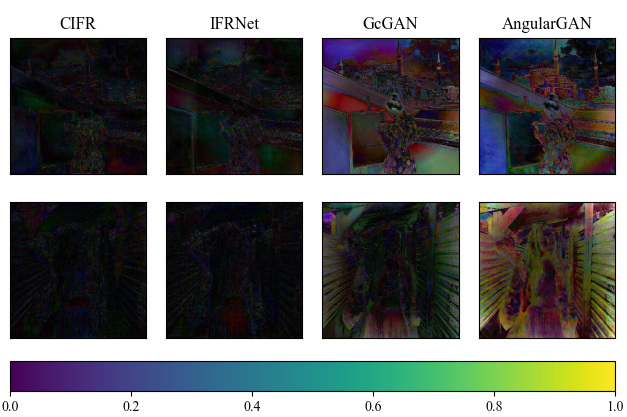}
        
        \caption{The residuals extracted by the absolute difference between the recovered images and their original version.}\label{fig:fig4} 
\end{figure}

We present the qualitative results of our proposed architecture and the other compared methods on Instagram filter removal in Figure \ref{fig:fig3}. When compared to the previous studies, the results show that CIFR improves the quality of recovered images by composing adaptive feature normalization idea and multi-layer patch-wise contrastive style learning for filter removal. CIFR minimizes the inconsistency on the background and foreground color tones, and it leads to have less artifacts (\ie checkerboard, false color, filter residuals) on different parts of the outputs. At this point, we present the comparison of the residual images of the most successful four methodologies in our benchmark in Figure \ref{fig:fig4}. This figure verifies that the residuals are typically formed by the visual effects brought by the corresponding filter, and CIFR performs best on effectively removing these effects. The residuals are calculated by the scaled absolute error between the output and the original version.

\subsection{Quantitative Analysis}

We have followed the same procedure in \cite{Kinli_2021_CVPR} for evaluating the quantitative performance of CIFR where four common image similarity metrics are employed in the experiments. These metrics are SSIM, PSNR, Learned Perceptual Image Patch Similarity (LPIPS) \cite{zhang2018perceptual} and CIE 2000 Color Difference (CIE-$\Delta$E) \cite{cie2000}. We show the results in Table \ref{tab:tab-1} for our proposed architecture and the other compared methods obtained by training on IFFI dataset \cite{Kinli_2021_CVPR}. Our method has generally better quantitative performance than the other methods in the benchmark. Particularly, CIFR surpasses the previous studies in SSIM and CIE-$\Delta$E metrics. For PSNR measurements, although CIFR outperforms the fundamental  \cite{pix2pix2017, CycleGAN2017} and the related \cite{FuCVPR19-GcGAN, DRIT_plus, park2020cut, sidorov2019conditional} image-to-image translation studies and the prominent method \cite{artisticFilter}, it falls behind IFRNet \cite{Kinli_2021_CVPR} by 0.02\%. 

\begin{figure*}[ht]
        \centering
        \begin{subfigure}{0.160\textwidth}
                \includegraphics[width=\textwidth]{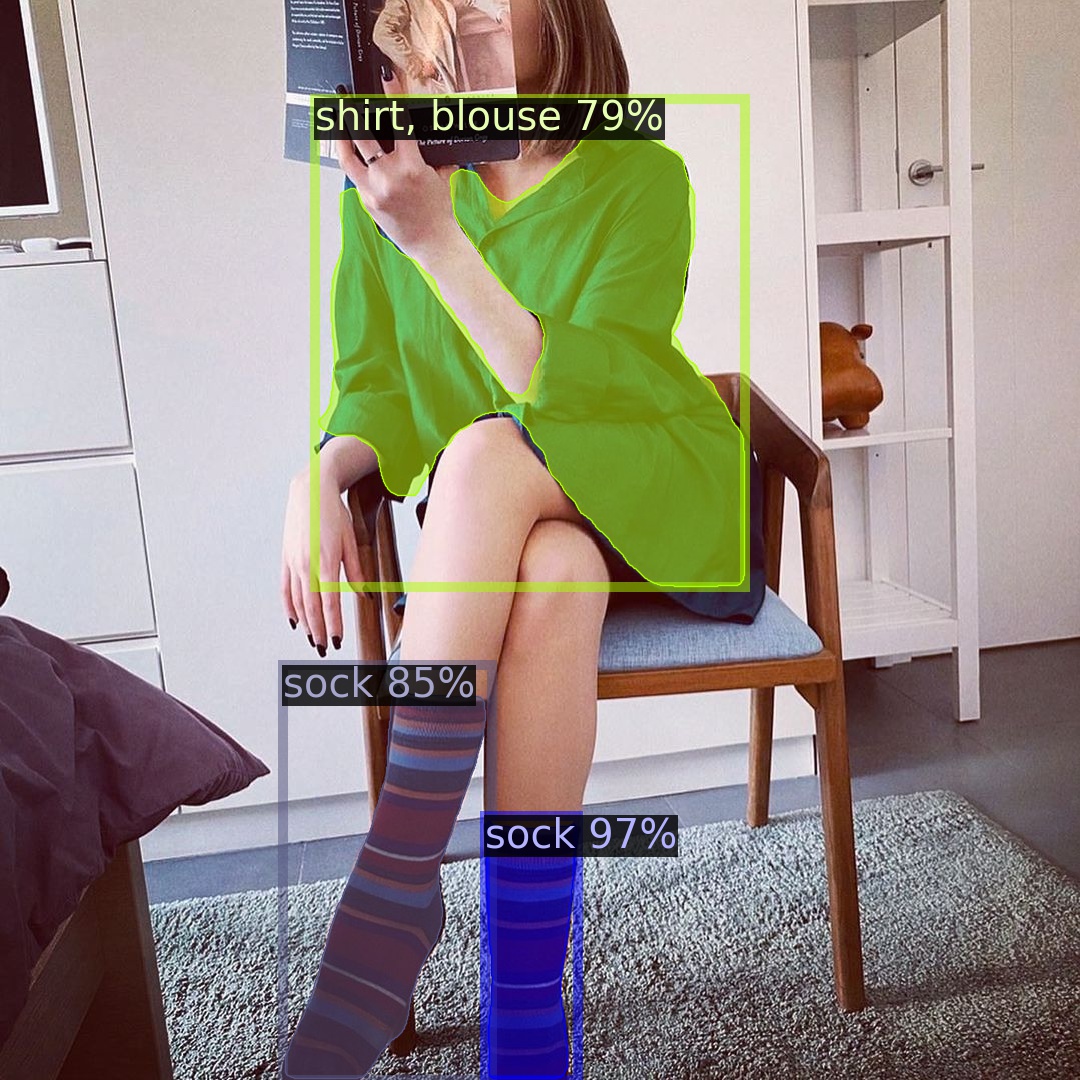}
                \includegraphics[width=\textwidth]{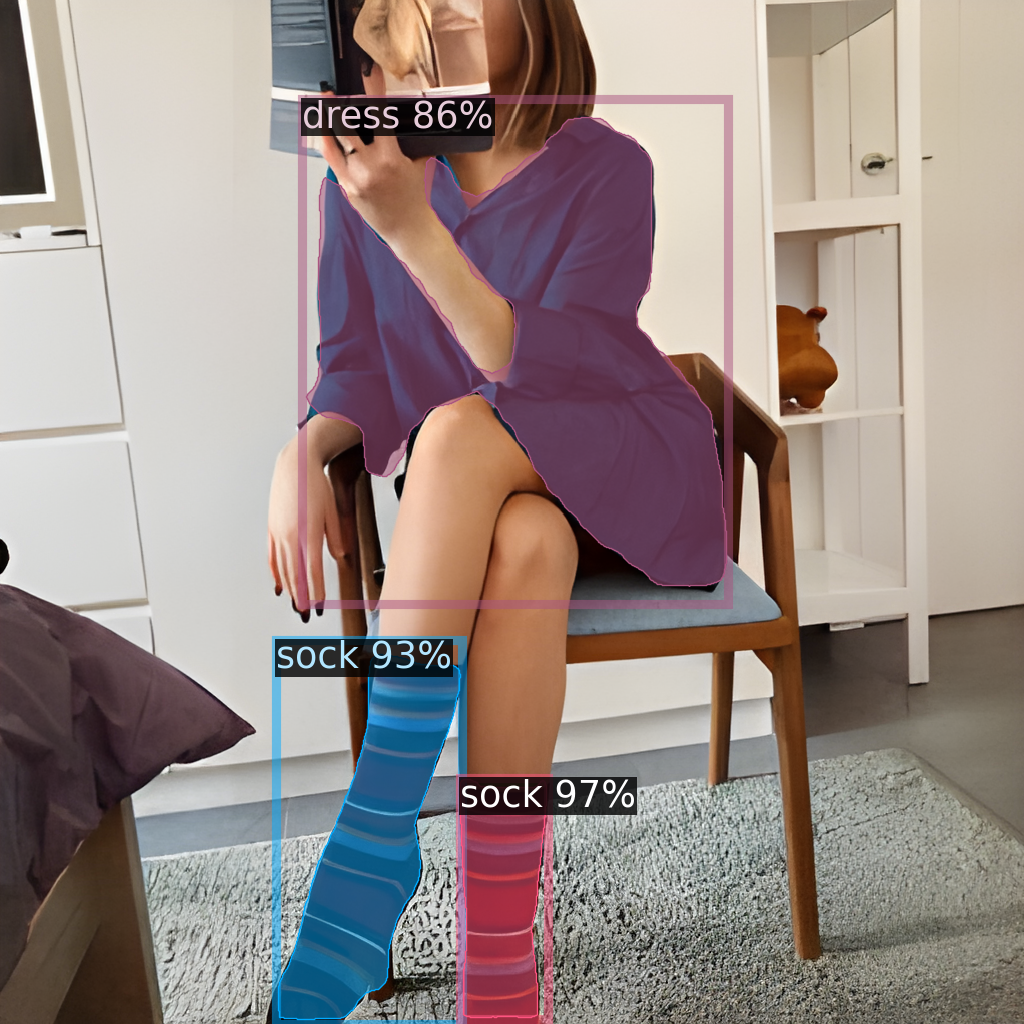}
                \caption{Hudson}
                \label{fig:fig5-hudson}
        \end{subfigure}       
        \begin{subfigure}{0.160\textwidth}
                \includegraphics[width=\textwidth]{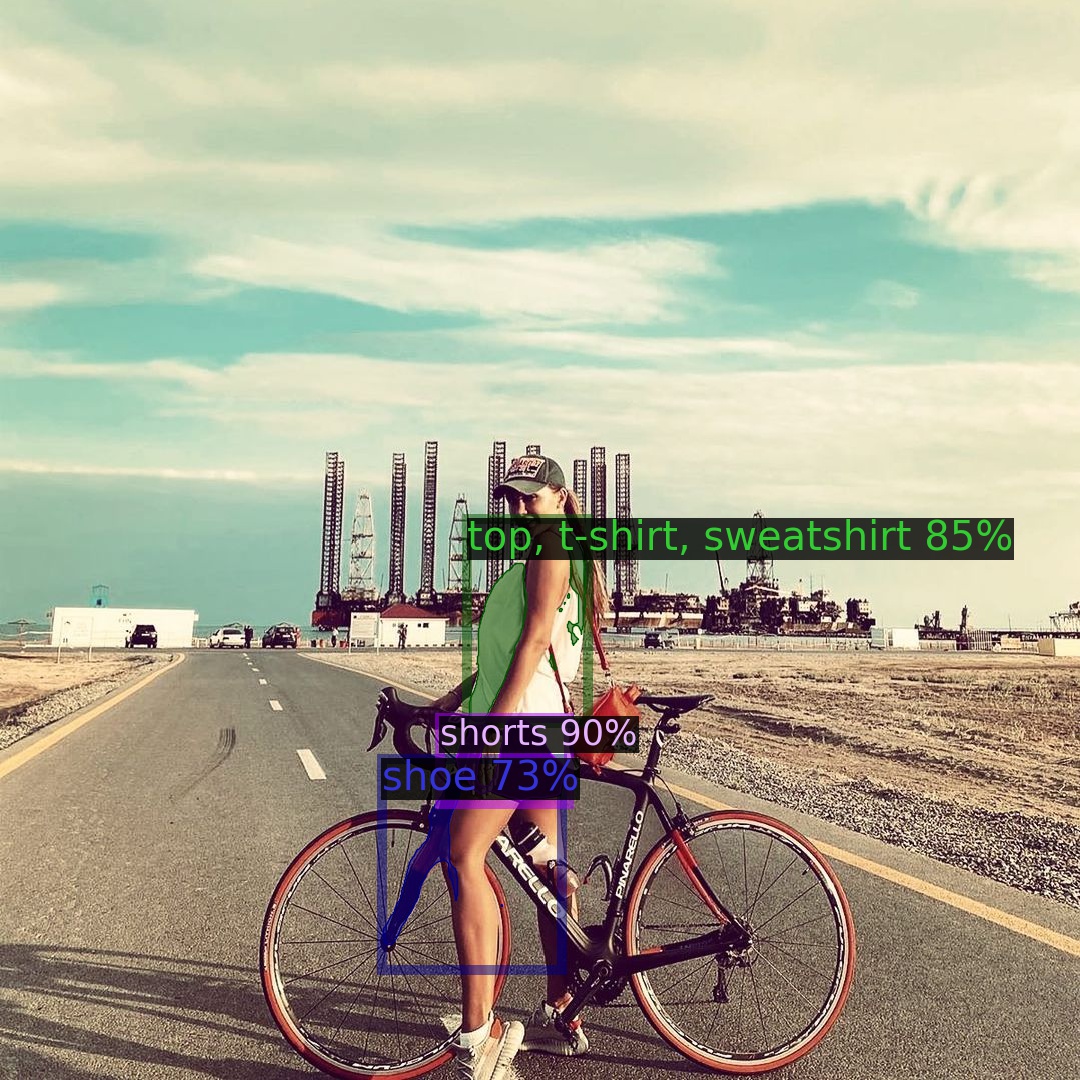}
                \includegraphics[width=\textwidth]{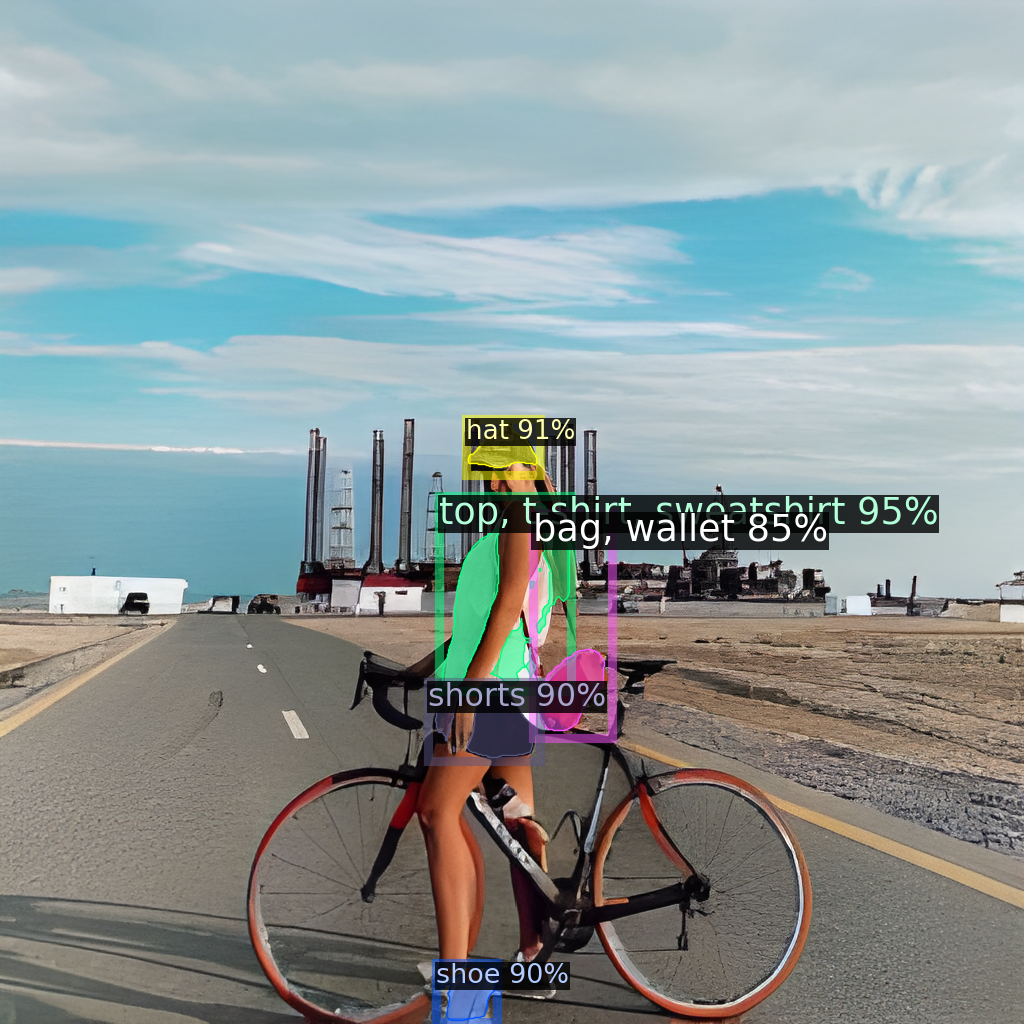}

                \caption{Brannan}
                \label{fig:fig5-brannan}
        \end{subfigure}
        \begin{subfigure}{0.160\textwidth}
                \includegraphics[width=\textwidth]{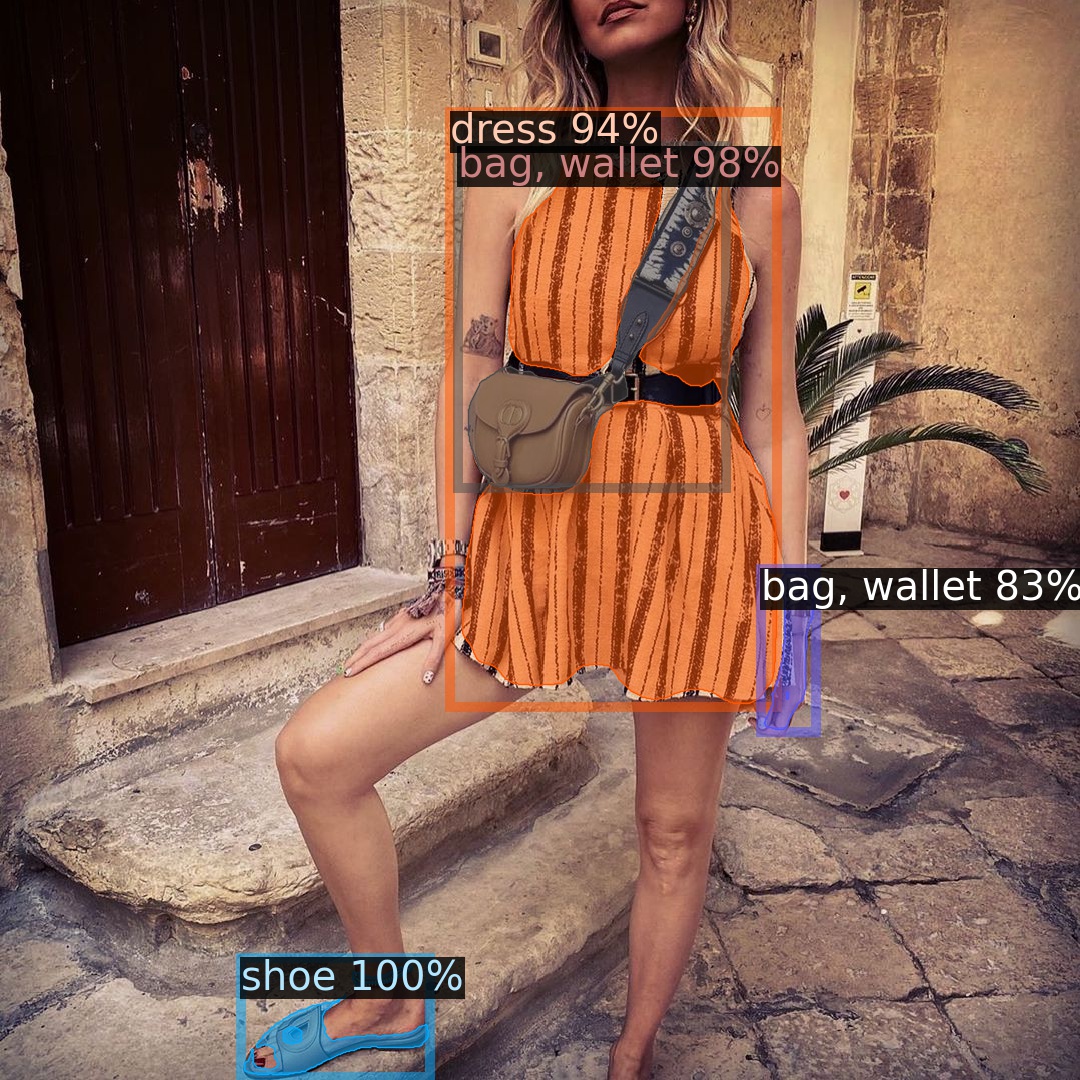}
                \includegraphics[width=\textwidth]{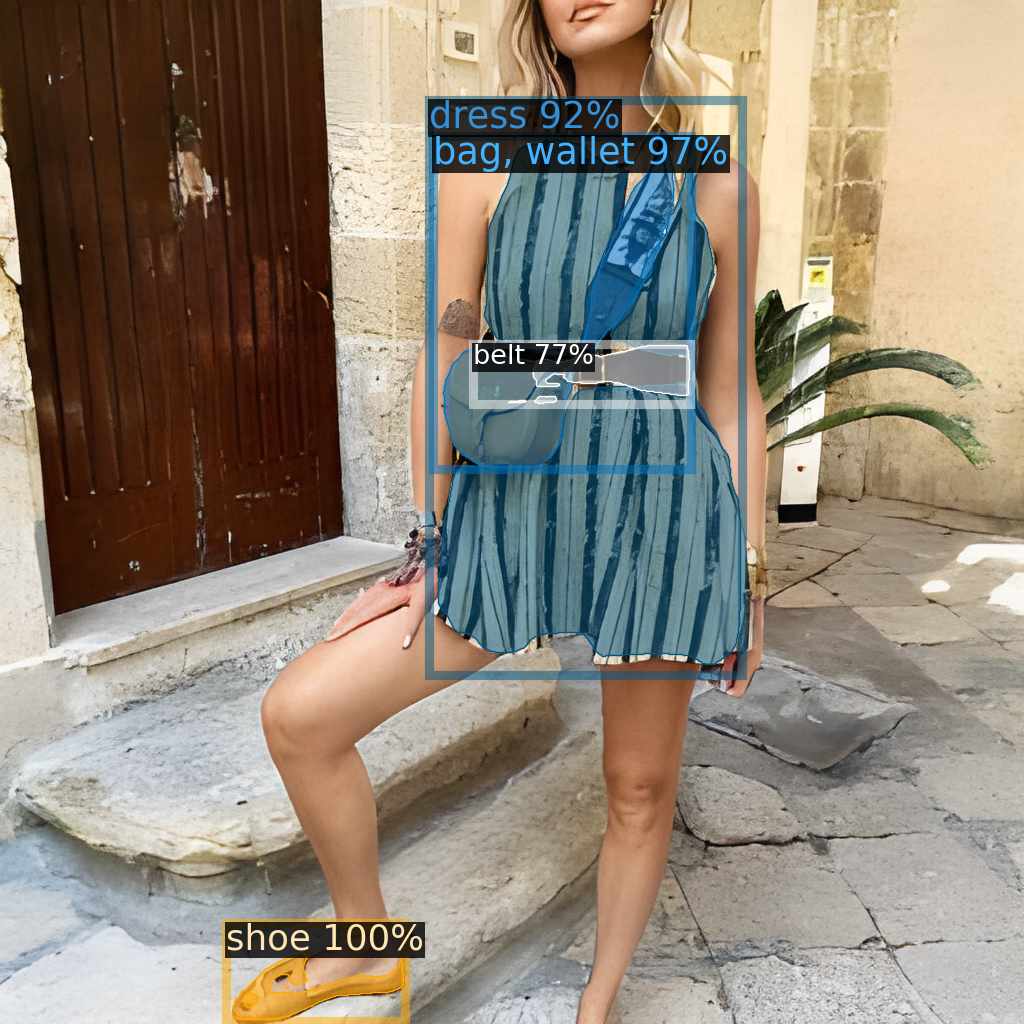}
                
                \caption{Sutro}
                \label{fig:fig5-sutro}
        \end{subfigure}
        \begin{subfigure}{0.160\textwidth}
                \includegraphics[width=\textwidth]{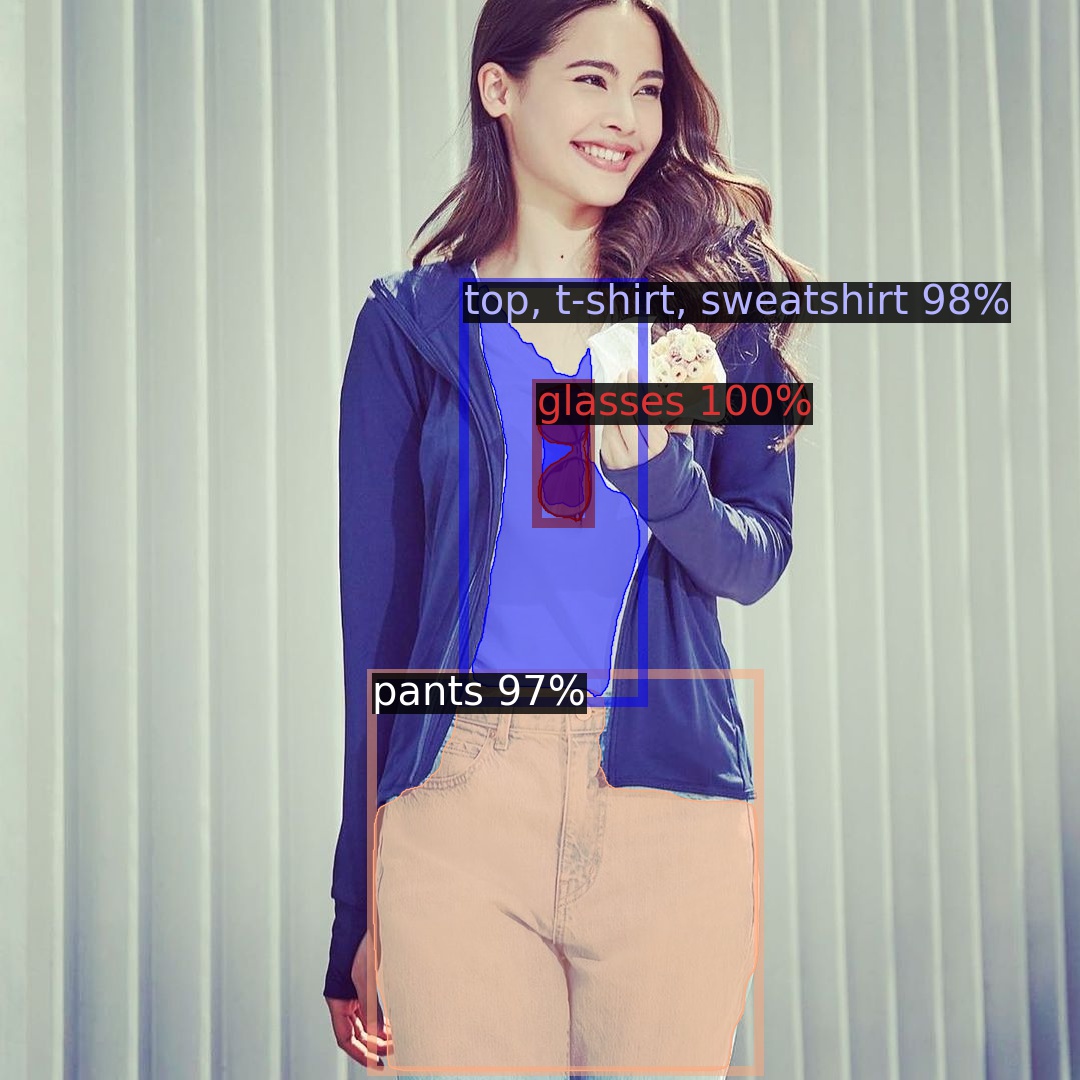}
                \includegraphics[width=\textwidth]{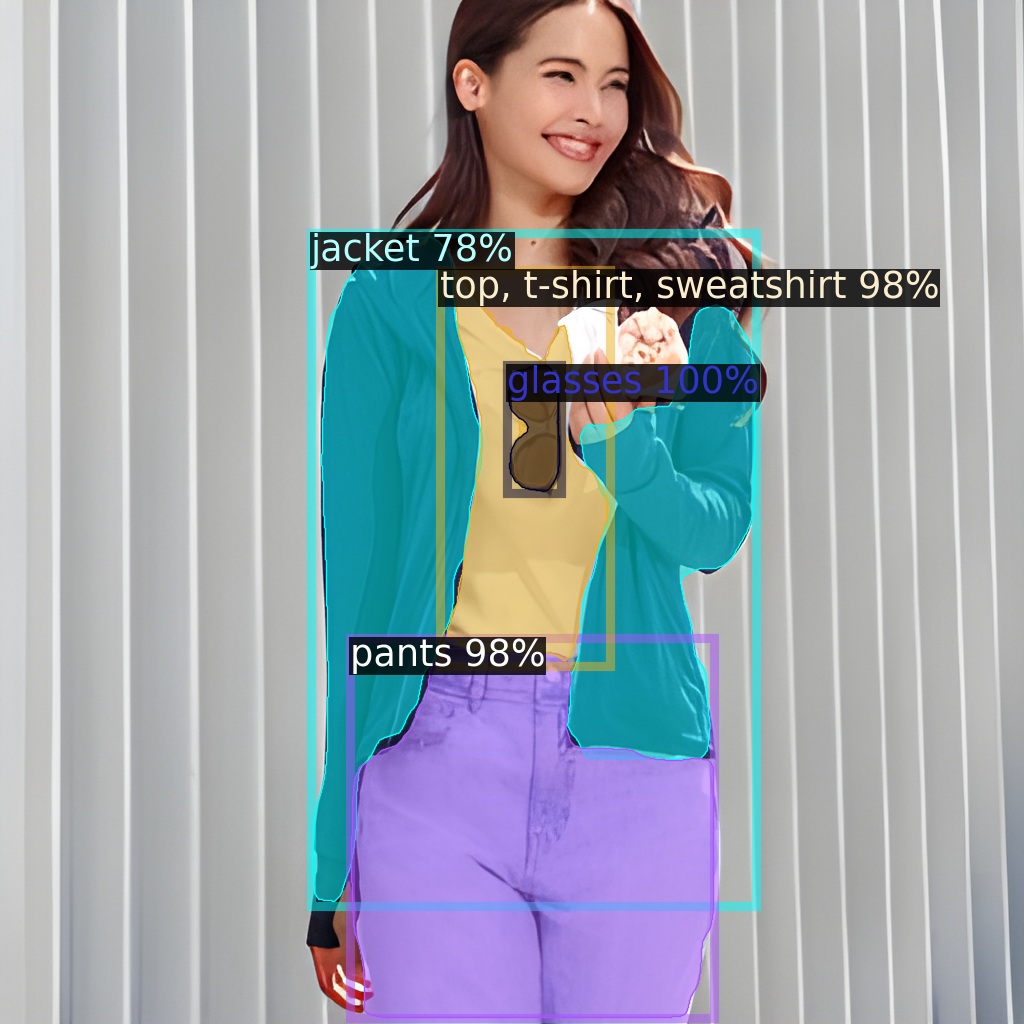}
                
                \caption{Amaro}
                \label{fig:fig5-amaro}
        \end{subfigure}
        \begin{subfigure}{0.160\textwidth}
                \includegraphics[width=\textwidth]{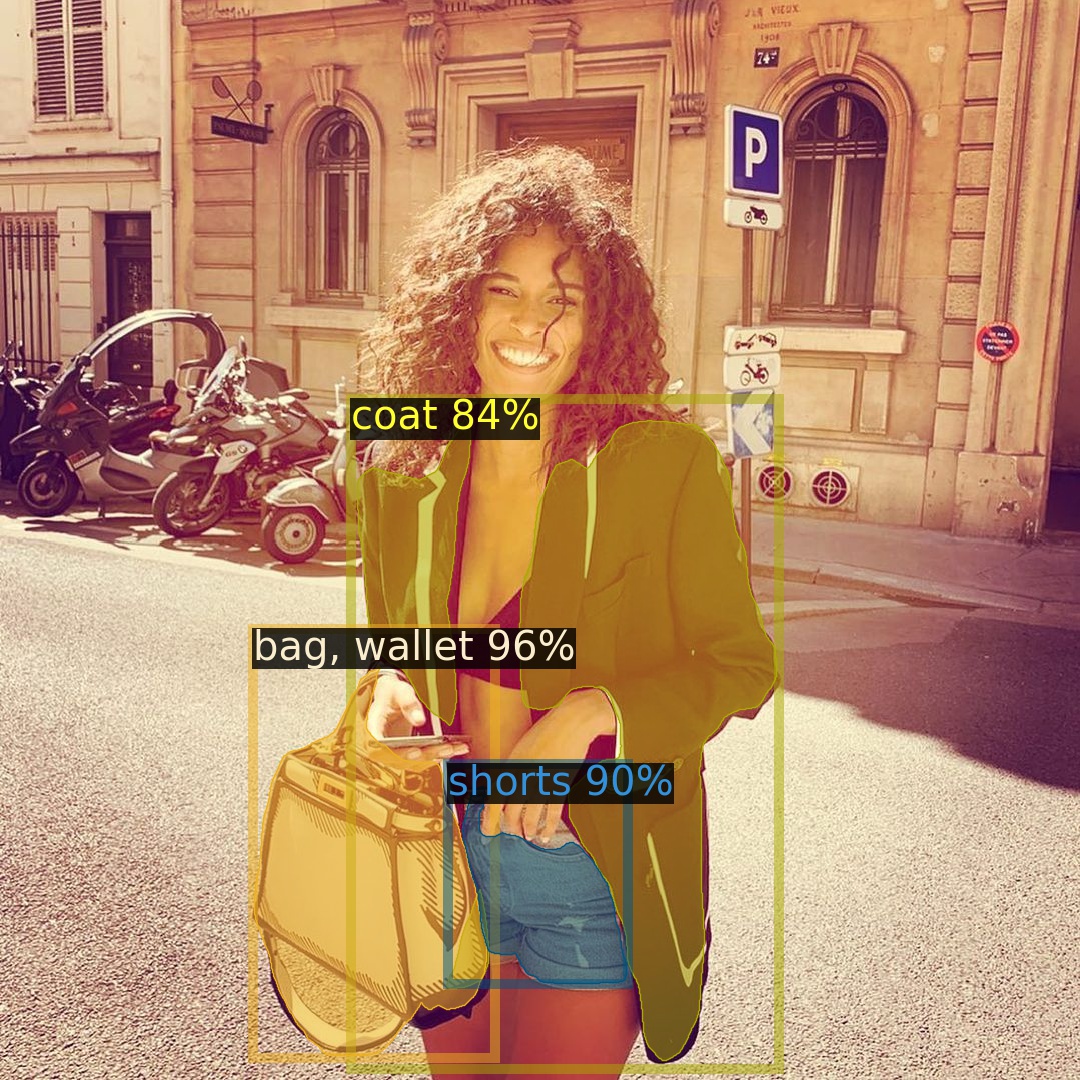}
                \includegraphics[width=\textwidth]{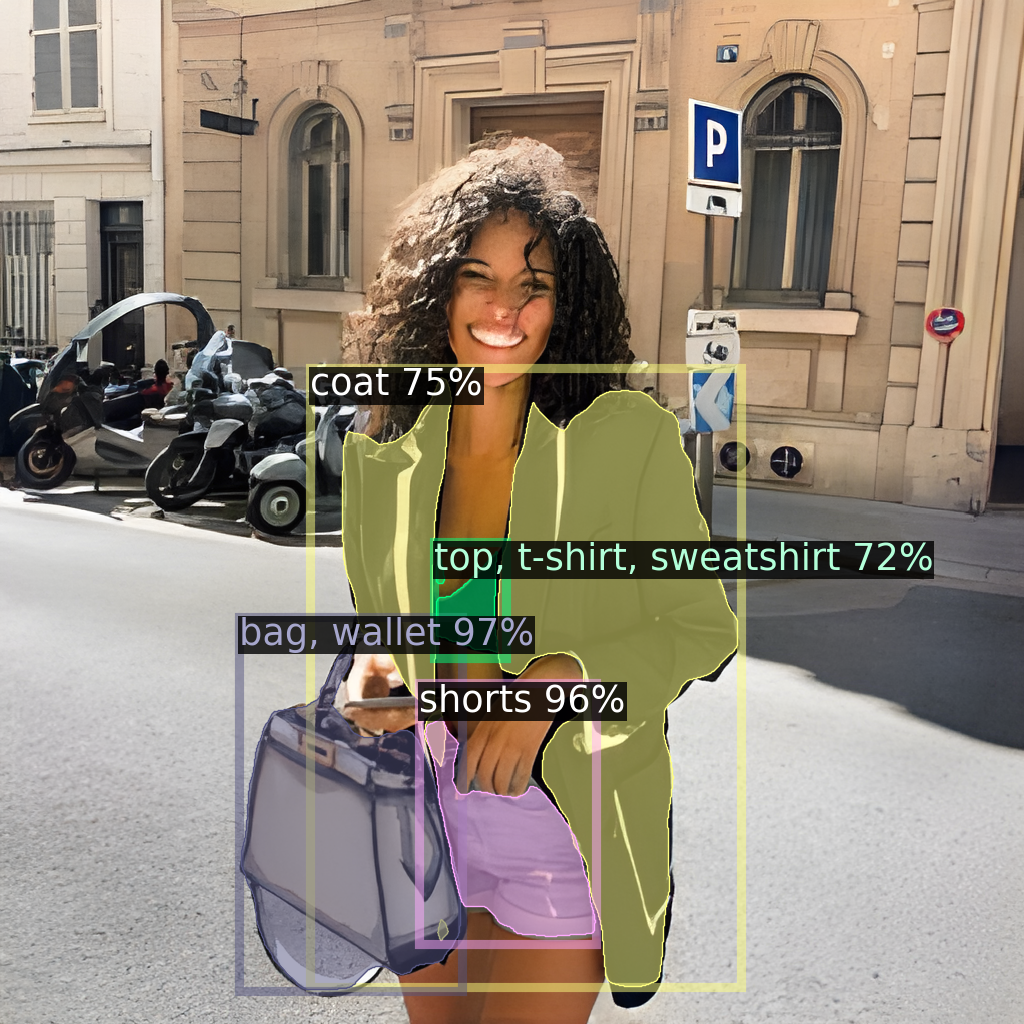}
                
                \caption{Toaster}
                \label{fig:fig5-toaster}
        \end{subfigure}
        \begin{subfigure}{0.160\textwidth}
                \includegraphics[width=\textwidth]{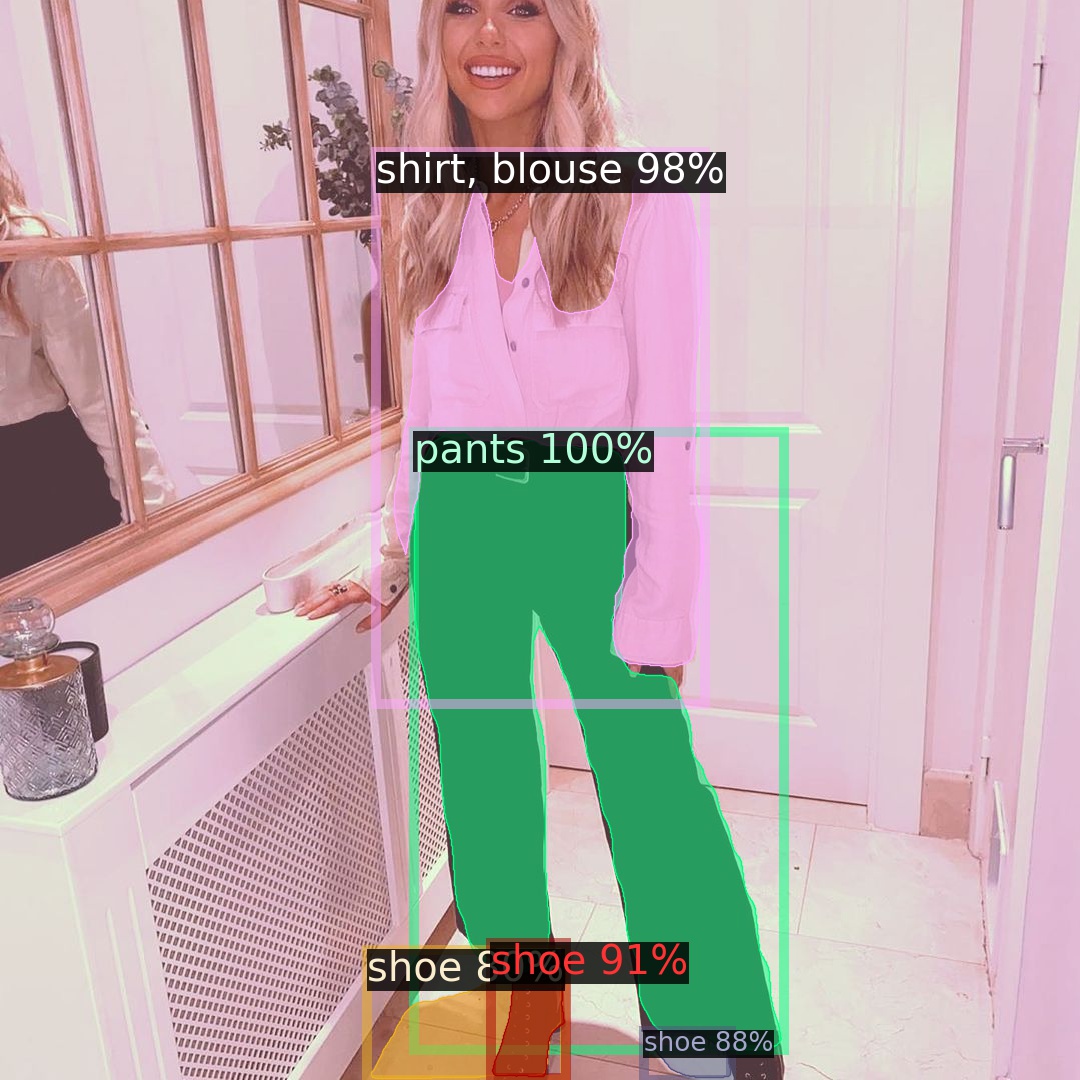}
                \includegraphics[width=\textwidth]{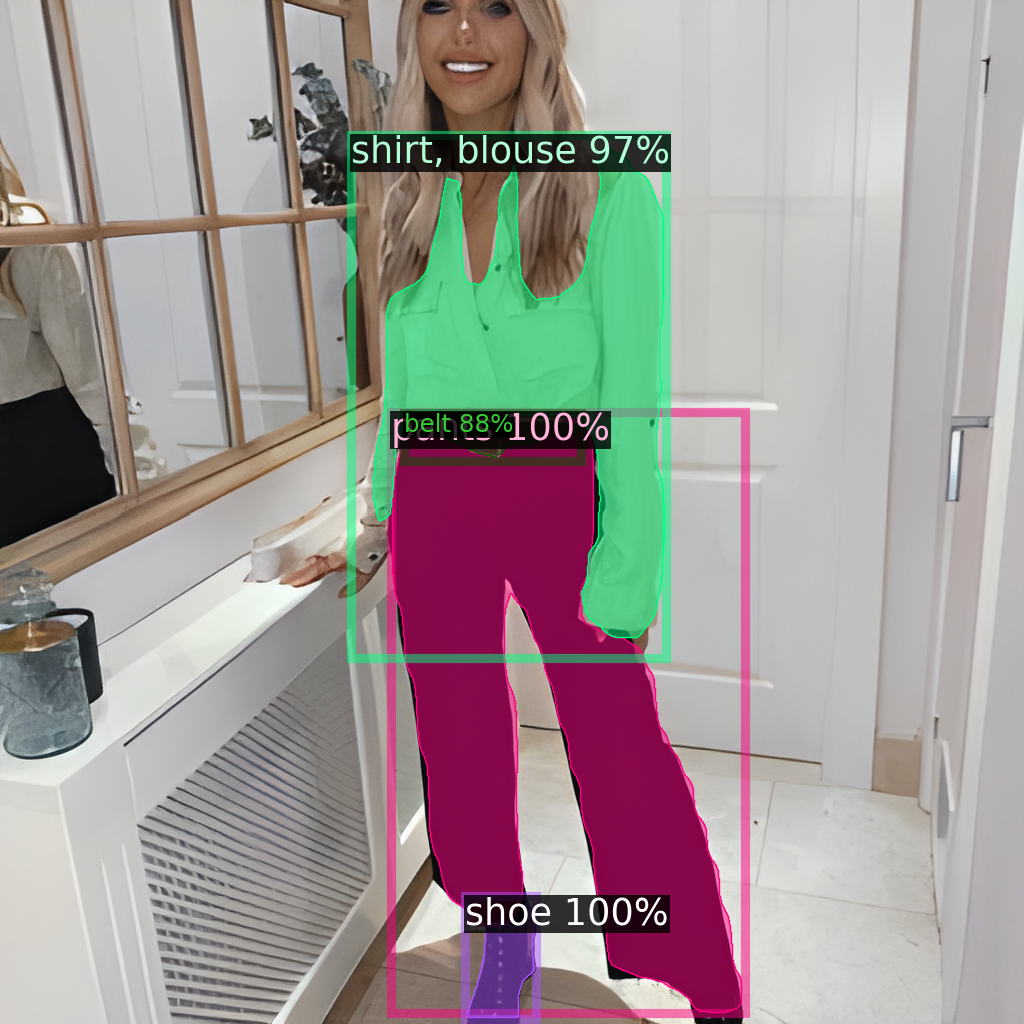}
                
                \caption{1977}
                \label{fig:fig5-1977}
        \end{subfigure}
        
        \caption{Demonstrating the impact of Instagram filter removal on downstream vision tasks like detection and segmentation. Examples are obtained from IFFI dataset \cite{Kinli_2021_CVPR}, and predicted by Attr-Mask-RCNN trained on Fashionpedia dataset, which is introduced in \cite{Jia2020Fashionpedia}. Rows: (1) the results of the images filtered by different Instagram filters, (2) the results of the images unfiltered by CIFR. Zoom in for better view.}\label{fig:fig5} 
\end{figure*}

\paragraph{Ablations} In this study, we build our proposed architecture in different settings to observe its performance in detail. The changes in our settings can be listed as follows: (1) we start our training from scratch, not using pre-trained weights of IFRNet. (2) we only include content PatchNCE loss to the objective functions as in \cite{park2020cut}. (3) we exclude Identity Regularization \cite{park2020cut} from the extended version of our PatchNCE loss. (4) we leave out the semantic and texture consistency losses used in \cite{Kinli_2021_CVPR}. Note that the exact same architecture, as shown in Figure \ref{fig:architecture}, and hyper-parameters are used for all these settings. Table \ref{tab:tab-1} also presents the results of additional experiments for proposed architecture. Our first observation is that we can achieve on-par performance by training our architecture from scratch, and it still performs better quantitative performance than the previous studies. Secondly, distilling the learning process of semantic and style similarities for the patch-wise contrastive learning strategy significantly improves the results of Instagram filter removal on all experiments. This demonstrates that capturing the pure style of the original images is one of the key aspects of removing the filters. Next, Identity Regularization has limited impact on the overall performance, and it can be omitted for reducing the training workload. Finally, excluding the consistency losses used in \cite{Kinli_2021_CVPR} from the final objective function leads to a decrease on the performance. However, these losses are quiet expensive functions for the training workload. Therefore, we believe that there is a still room for leaving out these expensive loss terms while improving the performance.

\begin{table}[h]
\begin{center}
\resizebox{\linewidth}{!}{\begin{tabular}{c|c|c|c|c}

\textbf{Method} & \textbf{SSIM} $\uparrow$& \textbf{PSNR} $\uparrow$& \textbf{LPIPS} $\downarrow$& \textbf{CIE-$\Delta$E} $\downarrow$ \\
\hline
PE \cite{artisticFilter} & 0.748 & 23.41 & 0.069 & 39.55 \\
pix2pix \cite{pix2pix2017} & 0.825 & 26.35 & 0.048 & 30.32 \\
CycleGAN \cite{CycleGAN2017} & 0.819 & 22.94 & 0.065 & 36.59 \\
AngularGAN \cite{sidorov2019conditional} & 0.846 & 26.30 & 0.048 & 31.11 \\
IFRNet \cite{Kinli_2021_CVPR} & 0.864 & \textbf{30.46} & 0.025 & 20.72 \\
DRIT++ \cite{DRIT_plus} & 0.626 & 16.23 & 0.162 & 47.95 \\
GcGAN \cite{FuCVPR19-GcGAN} & 0.838 & 21.75 & 0.060 & 38.54 \\
FastCUT \cite{park2020cut} & 0.763 & 20.08 & 0.083 & 39.86 \\
CUT \cite{park2020cut} & 0.744 & 20.96 & 0.081 & 38.64 \\
\hline
CIFR-no-pre-training & \textbf{0.888} & 29.24 & 0.02441 & 20.65 \\  
CIFR-no-style-nce & 0.859 & 28.13 & 0.03426 & 23.01 \\  
CIFR-no-id-reg & 0.879 & 29.40 & 0.02528 & 19.82 \\  
CIFR-no-consistency & 0.874 & 29.42 & 0.02708 & 21.23 \\  
CIFR & 0.880 & 30.02 & \textbf{0.02321} & \textbf{19.05} \\  
\end{tabular}}
\end{center}
\caption{Quantitative comparison of proposed architecture, its own variants and the compared methods on IFFI dataset. Obtained the available results from \cite{Kinli_2021_CVPR}, and re-trained the rest from scratch.}
\label{tab:tab-1}
\end{table}
\begin{table*}[!ht]
\centering
\resizebox{\textwidth}{!}{%
\begin{tabular}{c|c|cccccc|cccccc}
\multicolumn{2}{c|}{{\textbf{Filters}}} & 
  \multicolumn{6}{c|}{\textbf{Localization (mAP)}} &
  \multicolumn{6}{c}{\textbf{Segmentation (mAP)}} \\
\multicolumn{2}{c|}{} &
  \textbf{Top} &
  \textbf{Shirt} &
  \textbf{Pants} &
  \textbf{Dress} &
  \textbf{Shoe} &
  \textbf{Glasses} &
  \textbf{Top} &
  \textbf{Shirt} &
  \textbf{Pants} &
  \textbf{Dress} &
  \textbf{Shoe} &
  \textbf{Glasses} \\ \hline
{1977} &
  Filtered &
  7.976 & 0.000 & 11.348 & 5.406 & 16.084 & 9.505 & 
  9.773	& 0.000 &	10.228 &	6.713 &	13.424 &	7.657 \\
 &
  R-IFRNet &
  12.653 &	6.931 &	13.871	 &	11.042 &	24.318 &	\textbf{13.175}	 &
  11.521 &	7.178 &	12.815 &		11.978 &	\textbf{19.716} &	9.769 \\
 & R-CIFR &
  \textbf{13.115} & \textbf{10.891} & \textbf{15.175} & \textbf{11.314} &	\textbf{24.332} & 10.297	&
  \textbf{14.088} &	\textbf{10.561} &	\textbf{13.307} &	\textbf{12.866} &	19.004 & \textbf{9.901} \\ \hline
{Amaro} &
  Filtered &
  11.269 &	2.970 &	14.132 &	7.525 &	21.051 &	7.525  &
  10.414 &	3.960 &	13.508 &	10.179 &	15.323 &	7.525 \\
 &
  R-IFRNet &
  \textbf{13.035} &	6.188 &	13.890 &	10.144 &	26.027 &	10.594 &
  \textbf{11.658} &	7.426 &	14.001 &	11.560 &	20.232 &	9.208 \\
 & R-CIFR &
  12.673 &	\textbf{8.168} &	\textbf{16.006} &	\textbf{11.083 }&	\textbf{28.626} &	\textbf{10.693} &
  11.057 &	\textbf{8.911} & \textbf{14.598} & \textbf{11.644} &	\textbf{22.275} &	\textbf{9.901} \\ \hline
{Brannan} &
  Filtered &
  10.790 &	2.475 &	13.228 &	6.943 &	19.572 &	10.990 &
  11.607 &	3.960 &	11.484 &	7.017 &	15.271 &	7.168 \\
 &
  R-IFRNet &
  13.673 &	6.931 &	12.895 &	\textbf{10.562} &	\textbf{26.027} &	8.911 &
  13.359 &	6.436 &	12.665 &	\textbf{11.453} &	\textbf{21.615} &	8.020 \\
 & R-CIFR &
 \textbf{14.999} &	\textbf{9.901} &	\textbf{13.516} &	9.537 &	25.709 &	\textbf{11.221} &
  \textbf{15.200} &	\textbf{11.634} &	\textbf{13.264} &	10.911 &	20.977 &	\textbf{8.581} \\ \hline
{Hudson} &
  Filtered &
  13.294 & 5.941 & 13.512 & 9.285 & 24.554 & 13.861 & 
  13.818	& 5.941 &	12.437 &	11.243 &	18.329 &	10.693 \\
 &
  R-IFRNet &
  \textbf{15.093} &	6.188 &	13.964	 &	10.664 &	27.041 &	13.812	 &
  15.558 &	7.426 &	\textbf{14.420} &	\textbf{11.863} & 21.283 &	\textbf{11.023} \\
 & R-CIFR &
  14.322 & \textbf{10.297} & \textbf{14.844} & \textbf{10.673} &	\textbf{29.872} & \textbf{11.287}	&
  \textbf{15.815} &	\textbf{11.337} & 14.308 & 11.241 &	\textbf{21.654} & 10.297 \\ \hline
{Nashville} &
  Filtered &
  12.322	& 6.931	& 12.110	& 10.326 & 21.806 &	\textbf{11.089} & 
  11.432 & 	6.436 & 11.305 & 10.927 & 16.387 & 	8.079 \\
 &
  R-IFRNet &
  13.707 &	6.931 &	14.645 & \textbf{10.686}	& 24.811 &	7.525	 &
    14.546  &	7.426  &	12.643  & 10.485  &	19.994  &	6.733 \\
 & R-CIFR &
  \textbf{15.077} & \textbf{9.571} &	\textbf{15.705} &	9.884 &	\textbf{28.064} &	10.108	&
  \textbf{14.712} &	\textbf{9.901} &	\textbf{13.452} &	\textbf{11.193} &	\textbf{22.437} &	\textbf{8.515} \\ \hline
{Perpetua} &
  Filtered &
  14.494 &	5.941 &	14.238 &	7.475 &	21.133 &	\textbf{14.072} & 
  14.628 & 	5.941 & 	12.202 &  8.376 & 	17.263 & 	\textbf{12.208} \\
 &
  R-IFRNet &
  15.407 &	6.188 &	13.768 &	\textbf{11.634} &	26.154 &	12.541 	 &
  15.879 &	6.931 &	12.932	& \textbf{11.997} &	19.264 &	10.693 \\
 & R-CIFR &
  \textbf{16.903} &	\textbf{8.168} &	\textbf{15.880} &	11.541	& \textbf{28.047} &	13.333	&
  \textbf{16.939} & \textbf{9.406}	& \textbf{13.186} &	11.861	 & \textbf{22.065}	& 10.033 \\ \hline
{Valencia} &
  Filtered &
   12.481 &	6.188 &	12.105 &	9.010 &	23.083 &	9.901 & 
  12.558 &	7.426 &	10.671 &	9.036 &	18.131 &	7.683 \\
 &
  R-IFRNet &
  \textbf{14.490} &	6.436 &	\textbf{15.904} &	10.771 &	27.347 &	10.337	 &
  14.315 &	7.178 &	14.138 &	\textbf{12.291} &	20.624 &	9.743 \\
 & R-CIFR &
  14.467 &	\textbf{9.901} &	14.809 &	\textbf{11.735} &	\textbf{30.238} &	\textbf{10.693}	&
  \textbf{14.932} &	\textbf{9.653} &	\textbf{14.464} &	12.263 &	\textbf{23.358} &	\textbf{10.198} \\ \hline
{X-Pro II} &
  Filtered &
  13.604 &	6.188 &	12.555 &	8.465 &	21.637 &	\textbf{12.752} & 
  12.389 &	6.188 &	11.111 &	8.540 &	16.369 &	9.795 \\
 &
  R-IFRNet &
  \textbf{15.252} &	8.168 &	13.746 & 	10.815 &	25.722 &	11.116	 &
  15.751 &	8.911 &	12.605 &	12.546 &	19.757 &	9.003 \\
 & R-CIFR &
  15.189 &	\textbf{9.818} &	\textbf{14.538} &	\textbf{12.397} &	\textbf{27.538} &	12.488	&
  \textbf{16.253} &	\textbf{9.571} &	\textbf{13.360} &	\textbf{13.385}	& \textbf{21.394} &	\textbf{10.078} \\ \hline
{Original} &
  - &
  \color{red}\textbf{17.639} &	9.941 &	\color{red}\textbf{17.425} &	\color{red}\textbf{13.223} &	\color{red}\textbf{30.868} &	\color{red}\textbf{17.471} &
  \color{red}\textbf{18.758} &  10.178 &	\color{red}\textbf{16.432} &	\color{red}\textbf{15.509} &	\color{red}\textbf{24.937} &	\color{red}\textbf{15.278}
\end{tabular}%
}
\caption{Quantitative comparison for employing filter removal strategy to the data before feeding it to the downstream vision tasks. We present per-category mean average precision (mAP) scores of both filtered and recovered images on localization and segmentation tasks. We use our proposed architecture, namely \textit{CIFR} and the prior work \cite{Kinli_2021_CVPR}, for Instagram filter removal, and \textit{Attr-Mask-RCNN} architecture proposed in \cite{Jia2020Fashionpedia} for clothing localization and segmentation, which is trained on Fashionpedia dataset \cite{Jia2020Fashionpedia} and tested on IFFI dataset \cite{Kinli_2021_CVPR}.}
\label{tab:downstream}
\end{table*}

\paragraph{Impact on Vision Tasks}
As pointed out in \cite{Kinli_2021_CVPR}, due to different distractive factors, CNNs may not perform well for the real-world applications as much as in the standard benchmark studies. Noise or blurring in real-world scenarios or different transformations applied to the images can be the examples of these distractive factors. Likewise, Instagram filters transform the images into a different version whose feature maps change substantially. These changes arguably lead to the performance degradation on visual understanding tasks. At this point, we demonstrate the impact of removing Instagram filters from the images before feeding them to the downstream vision models. To achieve this, we make inferences of filtered and recovered images for localization and segmentation tasks. Filtered test images of IFFI dataset and their unfiltered versions by our proposed architecture are predicted by \textit{Attr-Mask-RCNN} model, which is proposed in \cite{Jia2020Fashionpedia}, and trained on Fashionpedia dataset \cite{Jia2020Fashionpedia}. Note that we made annotated the localization and segmentation ground truth of IFFI dataset by human annotators. Figure \ref{fig:fig5} shows that Instagram filters may degrade the performance of the downstream vision tasks (\eg misclassification, missing detections, wrong detection location, problematic segmentation maps, \textit{etc.}). This can be mitigated by wiping off the visual effects brought by these filters, and recover the images back to their original versions. In addition to this, we extend the evaluation of using filter removal strategy as a pre-processing step for the downstream vision tasks. In this experiment, we measure the performance of \textit{Attr-Mask-RCNN} on IFFI dataset by using per-category mean average precision (\textit{mAP}) of bounding boxes and segmentation masks. Table \ref{tab:downstream} presents the quantitative results of both filtered and recovered images on localization and segmentation tasks. Note that we pick 6 most frequently appeared clothing categories  of IFFI dataset for evaluating per-category mAP. The results support the main motivation behind the idea of directly removing any externally applied filters from the images to improve the overall performance of the downstream vision tasks. Moreover, it also verifies that different levels of perturbations have a negative impact on understanding the image contents, and thus hindering the further analysis of them in particular applications.

\section{Conclusions}
In this study, we propose a novel strategy for removing Instagram filters, which is a patch-wise contrastive learning mechanism for distilling the learning process of the semantic and style similarities. In addition to matching filtered and unfiltered patches at the same location, we also try to imitate the pure style of an original image by enabling a single negative instance for contrastive style learning, which has the same semantic information with the query instance, but with different style. Experiments show that our proposed architecture for this strategy mostly outperforms the performance of the previous studies on IFFI dataset and better to prevent to reduce the overall performance of the downstream vision models.

\paragraph{Discussion of Limitations}
We believe that there is still some room for improvements on the limitations of our strategy and the task itself. First, the volume and the diversity of the dataset used for this task (\ie the number of instances, filters and annotations) are limited when compared to the datasets of the other vision tasks, and it can be increased to be able to conduct more extensive experiments on this task. Next, the consistency loss shown as the part of Equation \ref{eq:final_loss} is an expensive and restrictive objective function for training workload. The hyper-parameters mainly defining the computational burden of training (\ie batch size and input size) mostly depend on this function. Therefore, a more elegant objective function can be designed for replacing with it. This architecture is implemented as a pre-processor for the downstream vision tasks, however it can be also designed as a single model where the filter removal is done just before the feature extraction, as in \cite{Talebi_2021_ICCV} for resizing.

{
    \small
    \bibliographystyle{ieee_fullname}
    \bibliography{macros,main}
}



\end{document}